\def\paperID{30907}
\def\confName{CVPR}
\def\confYear{2026}

\def\paperTitle{Voxify3D: Pixel Art Meets Volumetric Rendering}

\def\authorBlock{
    Yi-Chuan Huang
    \quad
    Jiewen Chan
    \quad
    Hao-Jen Chien
    \quad
    Yu-Lun Liu\vspace{0.5em}
    \\
    \centerline{National Yang Ming Chiao Tung University}
}

\newif\ifreview 
\newif\ifarxiv \newcommand{\arxiv}{\arxivtrue}
\newif\ifcamera 
\newif\ifrebuttal 
\arxiv 

\pdfoutput=1
\documentclass[10pt,twocolumn,letterpaper]{article}
\ifreview \usepackage[review]{cvpr} \fi
\ifarxiv \usepackage[pagenumbers]{cvpr} \fi
\ifrebuttal \usepackage[rebuttal]{cvpr} \fi
\ifcamera \usepackage{cvpr} \fi

\usepackage{graphicx}	
\usepackage{amsmath}	
\usepackage{amssymb}	
\usepackage{booktabs}
\usepackage{times}
\usepackage{microtype}
\usepackage{epsfig}
\usepackage{caption}
\usepackage{float}
\usepackage{placeins}
\usepackage{color, colortbl}
\usepackage{stfloats}
\usepackage{enumitem}
\usepackage{tabularx}
\usepackage{xstring}
\usepackage{multirow}
\usepackage{xspace}
\usepackage{url}
\usepackage{subcaption}
\usepackage{xcolor}
\usepackage[hang,flushmargin]{footmisc}
\usepackage{makecell}

\ifcamera \usepackage[accsupp]{axessibility} \fi

% ...

\ifarxiv  \fi

\newcommand{\R}[1]{{%
    \textbf{%
        \ifstrequal{#1}{1}{\textcolor{red}{R#1}}{%
        \ifstrequal{#1}{2}{\textcolor{blue}{R#1}}{%
        \ifstrequal{#1}{3}{\textcolor{magenta}{R#1}}{%
        \ifstrequal{#1}{4}{\textcolor{teal}{R#1}}{%
                           \textcolor{cyan}{R#1}%
        }}}}%
    }%
}}

\makeatletter
\renewcommand{\paragraph}{%
    \@startsection{paragraph}{4}%
    {\z@}{-0.5em}{-0.5em}%
    {\normalfont\normalsize\bfseries}%
}
\makeatother  

\usepackage{xr-hyper}

\makeatletter
\newcommand*{\addFileDependency}[1]{
  \typeout{(#1)}
  \@addtofilelist{#1}
  \IfFileExists{#1}{}{\typeout{No file #1.}}
}

\makeatother
\newcommand*{\myexternaldocument}[1]{
    \externaldocument{#1}
    \addFileDependency{#1.tex}
    \addFileDependency{#1.aux}
}

\definecolor{cvprblue}{rgb}{0.21,0.49,0.74}
\usepackage[pagebackref,breaklinks,colorlinks,allcolors=cvprblue]{hyperref}
\usepackage[capitalize]{cleveref}
\crefname{section}{Sec.}{Secs.}
\crefname{table}{Table}{Tables}
\crefname{figure}{Fig.}{Figs.}

\ifarxiv \crefname{appendix}{App.}{Apps.}
\else \crefname{appendix}{Suppl.}{Suppls.} \fi

\unless\ifarxiv \myexternaldocument{_supplementary} \fi

\begin{document}
\title{\paperTitle}
\author{\authorBlock}

\twocolumn[{%
\renewcommand\twocolumn[1][]{#1}%
\maketitle
\vspace{-7mm}
\begin{center}
\centering
\captionsetup{type=figure}
\resizebox{1.0\textwidth}{!} 
{
\includegraphics[width=\textwidth]{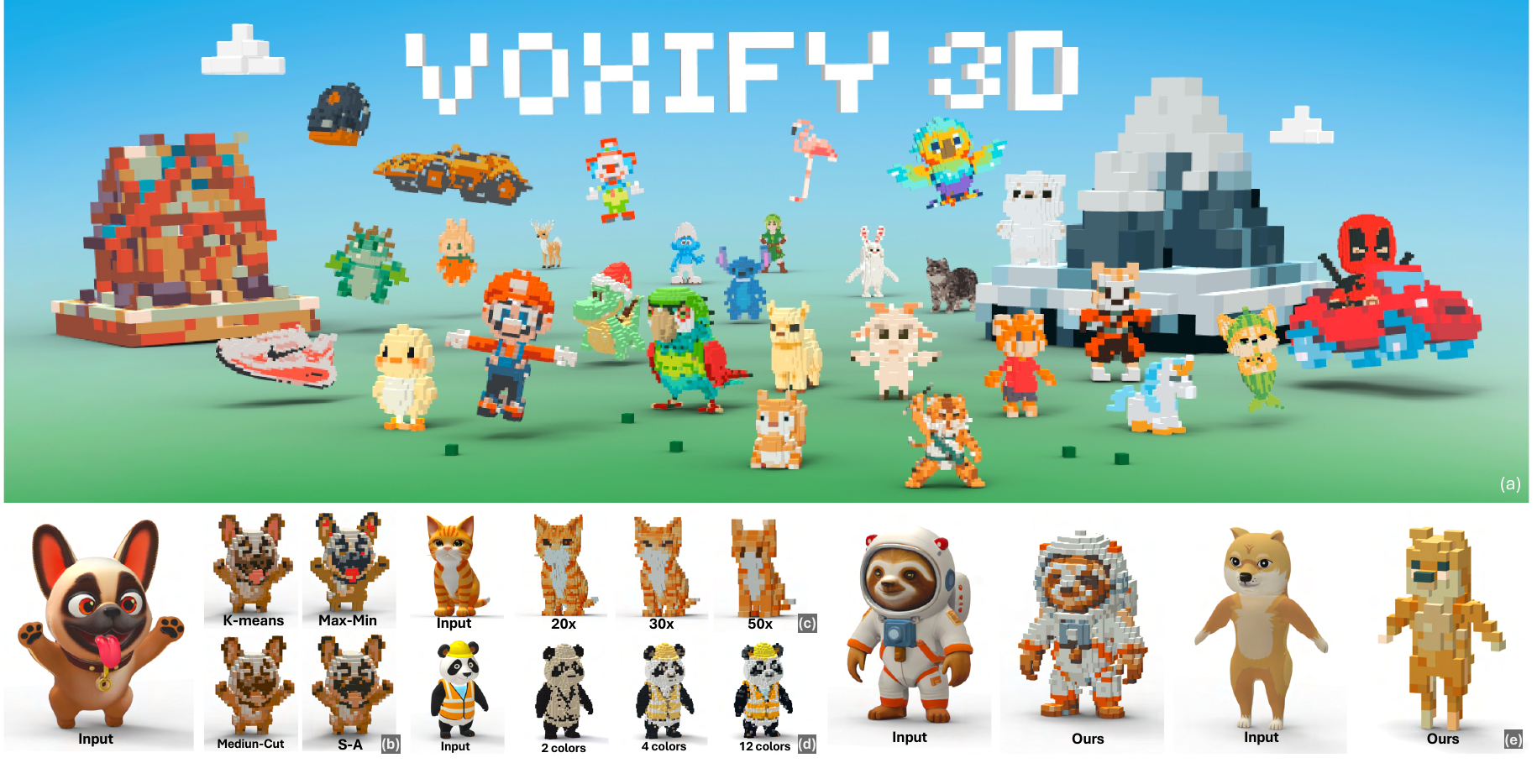}
}
\vspace{-7mm}
\caption{\textbf{Stylized voxel art with controllable abstraction.}
  Voxify3D converts 3D meshes into stylized voxel art using discrete color palettes, pixel art supervision, and voxel-based radiance fields. This teaser showcases the flexibility and quality of our method. (\textbf{a}) Diverse voxel art outputs across object types and use cases. (\textbf{b}) Comparison of different palette selection methods. (\textbf{c}) Control over the resolution of the voxel grid (20$\times$, 30$\times$, 50$\times$) allows a balance of detail and abstraction. (\textbf{d}) The variation in color count (2, 4, 8) shows the impact of palette size on expressiveness. (\textbf{e}) Input-output comparisons on multiple objects demonstrate faithful voxel structure, semantic clarity, and voxel art aesthetics.}
\label{teaser}
\end{center}
}]

\maketitle

\begin{abstract}
Voxel art is a distinctive stylization widely used in games and digital media, yet automated generation from 3D meshes remains challenging due to conflicting requirements of geometric abstraction, semantic preservation, and discrete color coherence. Existing methods either over-simplify geometry or fail to achieve the pixel-precise, palette-constrained aesthetics of voxel art. We introduce \textbf{Voxify3D}, a differentiable two-stage framework bridging 3D mesh optimization with 2D pixel art supervision. Our core innovation lies in the synergistic integration of three components: (1) \textbf{orthographic pixel art supervision} that eliminates perspective distortion for precise voxel-pixel alignment; (2) \textbf{patch-based CLIP alignment} that preserves semantics across discretization levels; (3) \textbf{palette-constrained Gumbel-Softmax quantization} enabling differentiable optimization over discrete color spaces with controllable palette strategies. This integration addresses fundamental challenges: semantic preservation under extreme discretization, pixel-art aesthetics through volumetric rendering, and end-to-end discrete optimization. Experiments show superior performance (37.12 CLIP-IQA, 77.90\% user preference) across diverse characters and controllable abstraction (2-8 colors, 20×-50× resolutions).
Project page: \url{https://yichuanh.github.io/Voxify-3D/}
\end{abstract}

\section{Introduction}

Voxel art is a distinctive form of 3D digital artwork, characterized by its minimalist aesthetic and discrete volumetric structure. Despite its growing popularity in games and digital media, creating high-quality voxel art remains challenging, requiring significant artistic expertise and manual effort. While recent works have achieved promising results in 2D pixel art stylization~\citep{wu2022make, han2018deep, binninger2024sdpixl, coutinho2022generating}, these techniques do not trivially extend to 3D voxel art. Directly using 2D pixel art for 3D reconstruction faces fundamental obstacles: projection-induced misalignment, multi-view inconsistencies, and ambiguous color representations.

\begin{figure}[t]
\centering
\small
  \setlength{\tabcolsep}{1pt}
\resizebox{\columnwidth}{!}{
  \begin{tabular}{ccc}
  \includegraphics[width=0.33\columnwidth]{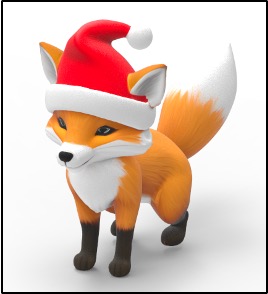} & 
  \includegraphics[width=0.33\columnwidth]{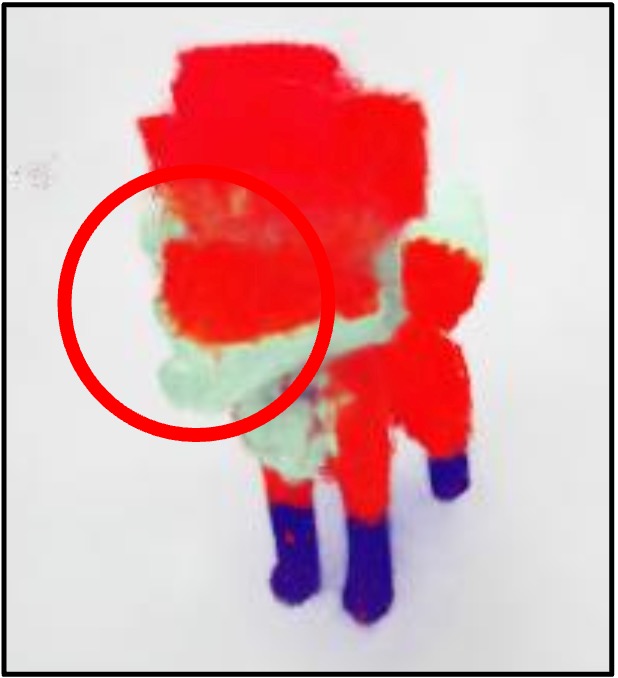} & 
  \includegraphics[width=0.33\columnwidth]{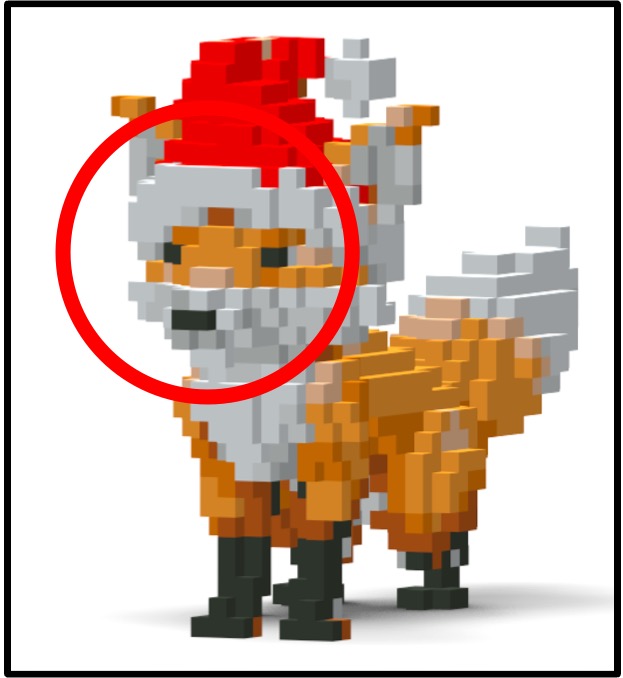} \\
  Input & Instruct-N2N~\cite{haque2023instruct} & Ours \\
  \includegraphics[width=0.33\columnwidth]{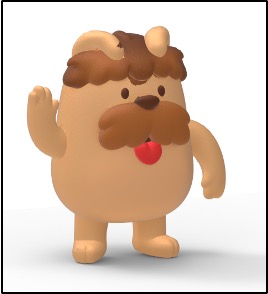} & 
  \includegraphics[width=0.33\columnwidth]{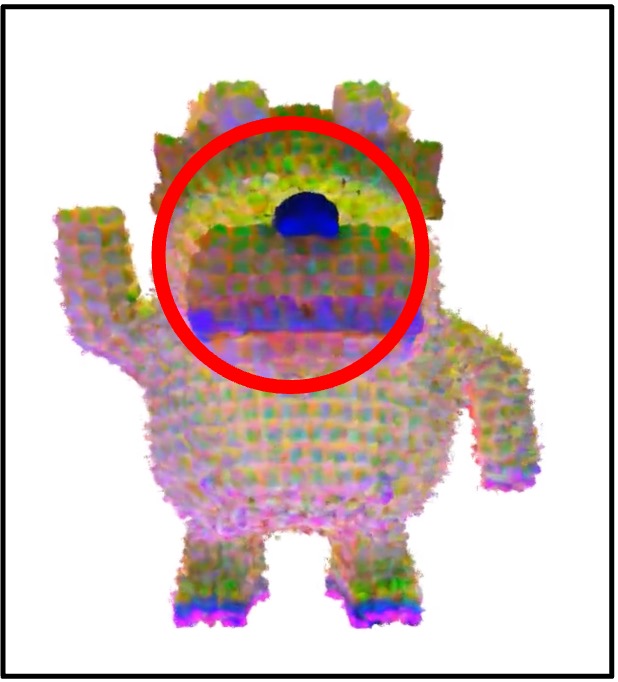} & 
  \includegraphics[width=0.33\columnwidth]{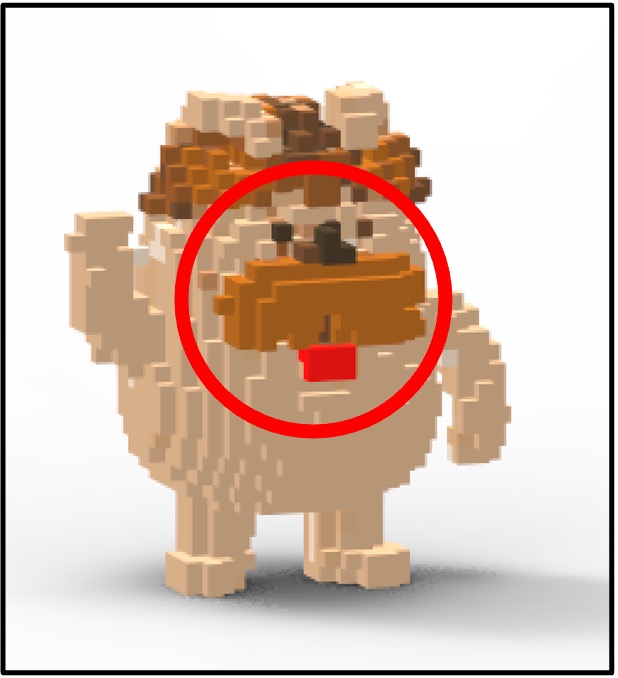} \\
  Input & Vox-E~\cite{sella2023vox} & Ours \\
  \includegraphics[width=0.33\columnwidth]{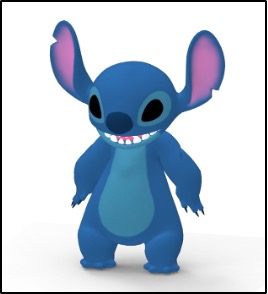} & 
  \includegraphics[width=0.33\columnwidth]{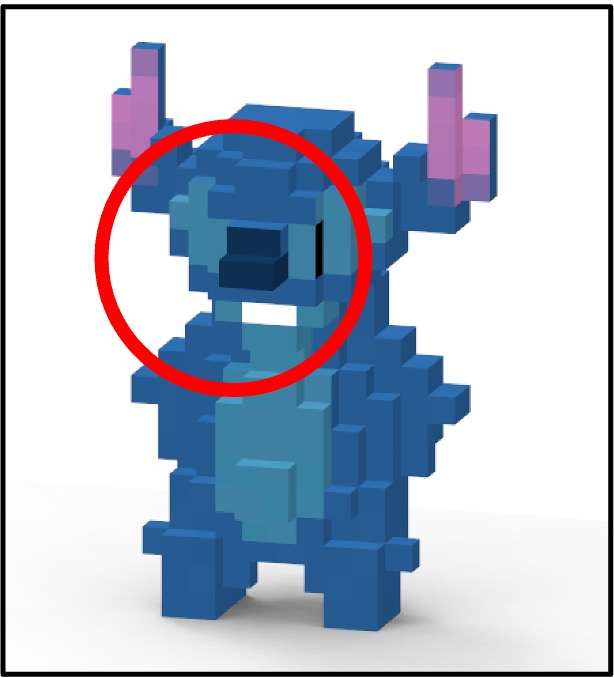} & 
  \includegraphics[width=0.33\columnwidth]{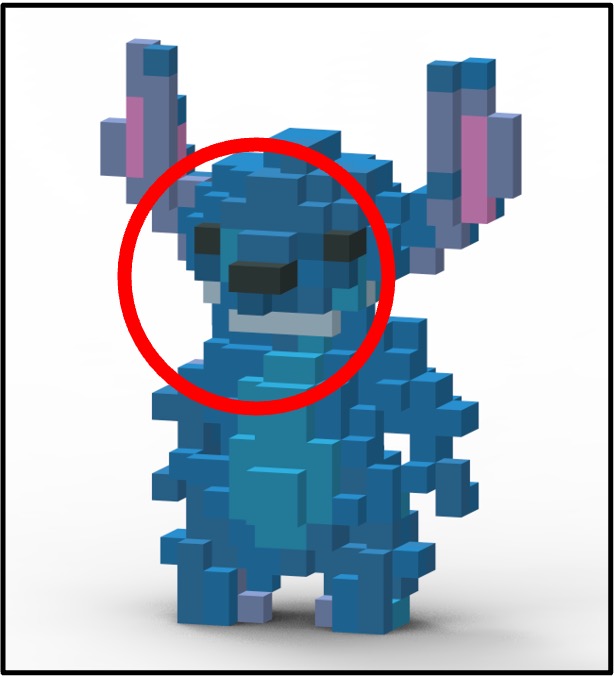} \\
  Input & Blender & Ours \\
  \end{tabular}
  }
\vspace{-3mm}
\caption{
\textbf{Existing methods often miss key features in voxelization.} While IN2N~\citep{haque2023instruct}, Vox-E~\citep{sella2023vox}, and Blender (Geometry Nodes) generate outputs that are coarse, blurry, or semantically inconsistent, they frequently lose critical elements such as facial features. In contrast, our method preserves structural details and produces visually appealing voxel art with sharp abstraction. 
}
\label{fig:motivaiton}
\end{figure}

Current voxel art generation from 3D meshes is limited. Simple downsampling loses semantic features, yielding overly coarse outputs. Voxel-based neural radiance fields~\citep{sun2022direct, chen2022tensorf, fridovich2022plenoxels} target photorealistic rendering, not stylistic abstraction. Neural editing methods~\citep{haque2023instruct, chen2023neuraleditor, clipnerf} struggle with clean, discrete representations. Procedural tools like Blender's Geometry Nodes require extensive manual tuning and lack unified optimization for discrete color control and semantic preservation, both critical for voxel art aesthetics. As~\cref{fig:motivaiton} shows, existing methods miss key features.

Voxel art generation poses three interrelated challenges that cannot be addressed by naively combining existing techniques: 
\textbf{(1) Alignment:} Perspective projection causes pixel-voxel misalignment, producing blurry gradients during optimization. Prior neural stylization~\cite{liu2023stylerf,wang2024nerfart} uses perspective rendering, unsuited for discrete art styles.
\textbf{(2) Semantic Preservation:} As resolution decreases, critical features (facial details, limb articulation) collapse. Standard perceptual losses on full images fail to capture local semantic importance.
\textbf{(3) Discrete Optimization:} Voxel art requires small palettes (2-8 colors), but gradient-based methods produce continuous values. Existing quantization~\cite{esser2021taming} lacks differentiability or user-controllable palette extraction.

We present \textbf{Voxify3D}, a principled framework addressing these challenges through synergistic technical design. We bridge 3D optimization with 2D pixel art supervision via: (1) six-view \textit{orthographic} rendering that eliminates perspective distortion for precise alignment; (2) patch-based CLIP loss adapted to preserve semantics across discretization levels; (3) palette-constrained Gumbel-Softmax enabling differentiable discrete optimization with flexible extraction strategies. This integration requires careful synchronization of rendering strategy, loss formulation, and quantization timing, not a simple combination.

Our two-stage pipeline first initializes coarse voxel geometry and color via neural volume rendering, then refines using orthographic pixel art supervision with semantic and discrete color constraints.
Our technical contributions include:
\begin{itemize}
\item \textbf{Orthographic pixel art supervision.} First framework to bridge 2D pixel art with 3D voxel optimization by eliminating perspective misalignment, enabling precise gradient flow for discrete stylization across six canonical views.

\item \textbf{Resolution-adaptive semantic preservation.} Patch-based CLIP formulation maintaining object identity under extreme discretization (20×-50×), addressing semantic collapse that standard perceptual losses fail to prevent.

\item \textbf{Palette-constrained differentiable quantization.} End-to-end optimization pipeline integrating Gumbel-Softmax with flexible palette extraction (4 strategies), temperature scheduling, and logit-based representation for controllable discrete color spaces (2-8 colors).
\end{itemize}



\section{Related Work}
\paragraph{3D Representations: From Pixels to Voxels.}
Pixel art generation evolved from interpolation~\citep{gerstner2013pixelated} and content-aware downscaling~\citep{kopf2013content, johnson2016perceptual} to deep learning: paired~\citep{isola2017pix2pix} and unsupervised translation~\citep{han2018deep, wu2022make}, GANs~\citep{coutinho2022generating, serpa2019towards}, diffusion~\citep{binninger2024sdpixl}, and vector methods~\citep{jain2023vectorfusion, xing2024svgdreamer, igarashi2022pixelartadaptation}. For 3D, voxel-based methods accelerate neural fields~\citep{mildenhall2021nerf, liu2023robust, meuleman2023progressively} through explicit grids~\citep{sun2022direct, yu2022plenoxels, chen2022tensorf, muller2022instant, garbin2021fastnerf, reiser2021kilonerf, schwarz2022voxgraf}, differentiable voxelization~\citep{luo2024differentiable}, unified frameworks~\citep{wu2024univoxel}, hierarchical structures~\citep{ren2024xcube}, sparse architectures~\citep{chen2023voxelnext}, and compression~\citep{li2023vqrf, zhan2025cat3dgs}. Multi-scale voxel representations~\citep{lin2025frugalnerf}, geometry-aware voxel features~\citep{tu2023imgeonet}, tensorial decomposition~\citep{cheng2024improving}, and MVS-based methods~\citep{su2024boostmvsnerfs} further enhance reconstruction quality. Voxels support geometry processing~\citep{coeurjolly2018regularization}, storage~\citep{museth2013vdb}, and simulation~\citep{losasso2004simulating}. Recent feed-forward generation achieves scale through structured latents~\citep{xiang2024structured}, hierarchical diffusion reaching 1024³~\citep{ren2024xcube}, cascaded point clouds~\cite{yushi2025gaussiananything}, transformers on voxelized shapes~\cite{mo2023dit}, and voxelized SDFs~\cite{li2023diffusion}. \emph{Unlike} 2D stylization \emph{or} 3D photorealism, we address \emph{discrete, palette-constrained} voxel art by bridging pixel art supervision with volumetric optimization via orthographic alignment, extending voxel radiance fields~\citep{sun2022direct} with palette quantization.

\paragraph{Stylization and Discrete Color Control.}
Neural 3D stylization progressed from score distillation~\citep{poole2022dreamfusion} and CLIP guidance~\citep{michel2022text2mesh} to zero-shot transfer~\citep{liu2023stylerf}, painterly rendering~\citep{wang2024nerfart, sun2024stylizednerf, zhang2023refnpr, nguyen2024stylenerf2nerf}, high-resolution generation~\citep{lin2023magic3d, chen2023fantasia3d}, and \emph{local} control~\cite{decatur20243d, liu2025wir3d, gomes2024controllable, chen2024stylecity}. Gumbel-Softmax~\citep{jang2017categorical,maddison2017concrete, shah2024decoupled} enables discrete optimization in NAS~\citep{liu2018darts, cai2018proxylessnas}, VQ-VAE~\citep{oord2017vqvae, takida2024hqvae}, and neural fields~\citep{liu2024contentaware, chen2025qdit}. Score-based generative models~\citep{chao2022denoising} provide conditional generation through likelihood matching. Palette methods include 2D quantization~\citep{binninger2024sdpixl}, 3D color decomposition~\cite{kuang2023palettenerf}, material extraction~\cite{lopes2024material}, vector quantization~\cite{huang2023quantart}, and interactive editing~\cite{lee2023ice}, with alternatives like VQGAN~\citep{esser2021taming} and latent upsampling~\citep{menon2020pulse}. \emph{In contrast to} \emph{continuous} stylization and \emph{fixed} codebooks, we integrate Gumbel-Softmax with \emph{user-controllable} palette extraction (K-means, Max-Min, Median Cut, Simulated Annealing), synchronized scheduling, and logit-based representation for \emph{pixel-precise} voxel art.

\paragraph{Multi-view Supervision and Semantic Preservation.}
Multi-view consistency uses RL refinement~\citep{xie2024carve3d}, view aggregation~\citep{yang2023consistnet, su2024boostmvsnerfs}, and latent diffusion~\citep{voleti2024sv3d}. Orthographic projection serves specialized domains: aerial orthophotos~\cite{yue2025nerfortho, chen2024orthonerf}, CAD reconstruction~\cite{zhou2025gaussiancad}, and furniture assembly~\cite{hu2023plankassembly}. CLIP~\citep{radford2021learning} enables semantic guidance~\citep{frans2022clipdraw, wang2022clipnerf, tang2024dreamgaussian, li2023blendeddiffusion, kim2022diffusionclip, mokady2022clipcap, patashnik2021styleclip, chen2024clipdriven}, with text supervision extending to semantic segmentation~\citep{wu2024imagetext}. Semantic preservation under discretization uses masked autoencoders~\cite{li2023voxformer}, context-aware transformers~\cite{yu2024context}, semantic structures~\cite{li2024svdtree}, geometry-aware downsampling~\cite{pentapati2025geoscaler}, and hierarchical upsampling~\cite{ren2024scube}. \emph{Unlike} perspective stylization \emph{or} orthographic reconstruction, we combine orthographic rendering with pixel art supervision, designing \emph{resolution-adaptive patch-based} CLIP loss preventing semantic collapse at 20×-50× discretization where image-level losses fail.

\paragraph{Applications and Datasets.}
Mesh generation exploits diffusion and sparse views~\citep{wang2018pixel2mesh, liu2024one, bala2024edify, lin2023magic3d, xu2024instantmesh, hong2023lrm, wang2024prolificdreamer, huang2025spar3d, xiang2024structured, li2024genrc}, with character datasets~\citep{wang2022rodin, wu2024unique3d}. Game assets require structural decomposition~\cite{huang2025part}, PBR materials~\cite{zhang2024clay}, and procedural libraries~\cite{pyarelal2021modular, kim2024minecraft}. Fabrication includes LEGO generation~\citep{anonymous2025legogpt, ge2024lego}, Earth voxelization~\citep{lewis2024voxelizing}, and 3D printing~\citep{swaminathan2018voxelprinting}. These inform our evaluation but don't address \emph{mesh-to-voxel-art} conversion with \emph{semantic fidelity}, \emph{palette constraints}, and \emph{controllable} abstraction.

\section{Method}

\begin{figure*}[t]
\centering
\includegraphics[width=1.0\textwidth]{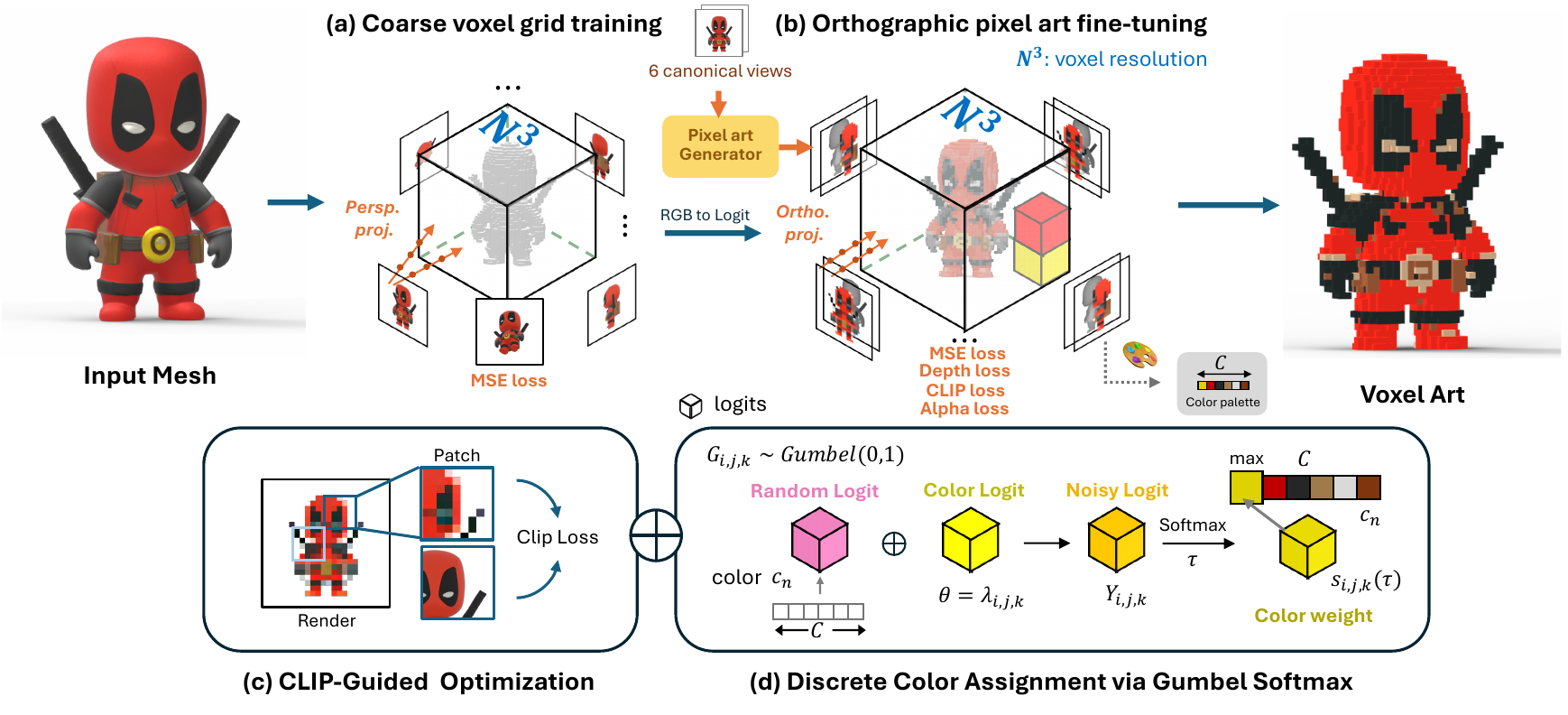}
\vspace{-3mm}
\caption{\textbf{Our two-stage voxel art generation pipeline.} 
(a) \textit{Coarse voxel grid training:} Given a 3D mesh, we render multi-view images and optimize a voxel-based radiance field (DVGO~\citep{sun2022direct}) using MSE loss to learn coarse RGB and density.  
(b) \textit{Orthographic pixel art fine-tuning:} We refine the voxel grid using six orthographic pixel art views, which also serve to extract a discrete color palette (e.g., via k-means). Optimization includes appearance, depth, and alpha losses.
(c) \textit{CLIP-guided optimization:} A CLIP loss computed over rendered patches and mesh images encourages semantic alignment while being memory-efficient.  
(d) \textit{Differentiable discrete color selection via Gumbel-Softmax:} Each voxel stores palette logits. Gumbel-Softmax enables differentiable sampling for end-to-end color optimization, yielding coherent, stylized voxel art.}
\label{fig:pipeline}
\end{figure*}

We propose a two-stage framework for converting 3D meshes into stylized voxel art with high fidelity and semantic consistency (Fig.~\ref{fig:pipeline}). 
Stage~1 (Sec.~\ref{sec:stage_1}) builds a coarse voxel radiance field using DVGO~\citep{sun2022direct} to establish geometric and color foundations. 
Stage~2 (Sec.~\ref{sec:stage_2}) refines the grid under orthographic pixel-art supervision, with CLIP-based loss (Sec.~\ref{sec:clip}) for semantic alignment and depth loss for geometric preservation. 
To achieve clean abstraction and a coherent palette, we replace the RGB grid with a learned color-logit grid and apply Gumbel-Softmax for differentiable palette quantization (Sec.~\ref{sec:gumbel}). 
This pipeline retains abstract details, enforces a dominant palette, and conveys the distinctive style of voxel art across resolutions.

\subsection{Coarse Voxel Grid Training} \label{sec:stage_1}
The first stage adapts DVGO~\citep{sun2022direct} to build a coarse voxel representation. 
Unlike NeRFs using MLPs, DVGO directly optimizes two explicit voxel grids: a density grid $d$ for spatial occupancy and a color grid $\mathbf{c}=(r,g,b)$ for appearance. 
This explicit structure enables faster training and efficient rendering.

We partition the object's bounding box into a grid of resolution $(W/\texttt{cell\_size})^3$, 
where $W$ is the canonical orthographic image width (pixels) and \texttt{cell\_size} is the number of pixels per voxel edge. Each voxel stores density $d$ and RGB color $\mathbf{c}$. The rendered color $C(\mathbf{r})$ along a camera ray $\mathbf{r}$ is computed as:
\begin{multline}
C(\mathbf{r}) = \sum_{k=1}^N T_k \alpha_k \mathbf{c}_k,
\quad
T_k = \exp\left(-\sum_{j=1}^{k-1} d_j \delta_j\right),\\
\alpha_k = 1 - \exp(-d_k \delta_k),\quad\quad\quad\quad\quad
\label{eq:vol_render}
\end{multline}
where $N$ is the number of samples along the ray, $d_k$ the density, $\delta_k$ the distance between consecutive samples, $T_k$ the accumulated transmittance, and $\alpha_k$ the opacity at sample $k$.

The coarse voxel grid is optimized with:
\begin{equation}
\mathcal{L}_\text{total} = 
\mathcal{L}_\text{render} + 
\lambda_d \mathcal{L}_\text{density} + 
\lambda_b \mathcal{L}_\text{bg},
\end{equation}
where $\mathcal{L}_\text{render}$ minimizes the MSE between rendered and target colors to ensure visual fidelity, 
$\mathcal{L}_\text{density}$ regularizes the density to suppress noise, prevent near-clip artifacts, and employs total variation (TV) regularization to enforce spatial smoothness, 
and $\mathcal{L}_\text{bg}$ uses entropy loss to maintain clear geometry and reduce background artifacts. 
This stage provides a good initialization for color and density.

\subsection{Orthographic Pixel Art Fine-tuning} \label{sec:stage_2}

\begin{figure}[t]
\centering
\includegraphics[width=1.0\columnwidth]{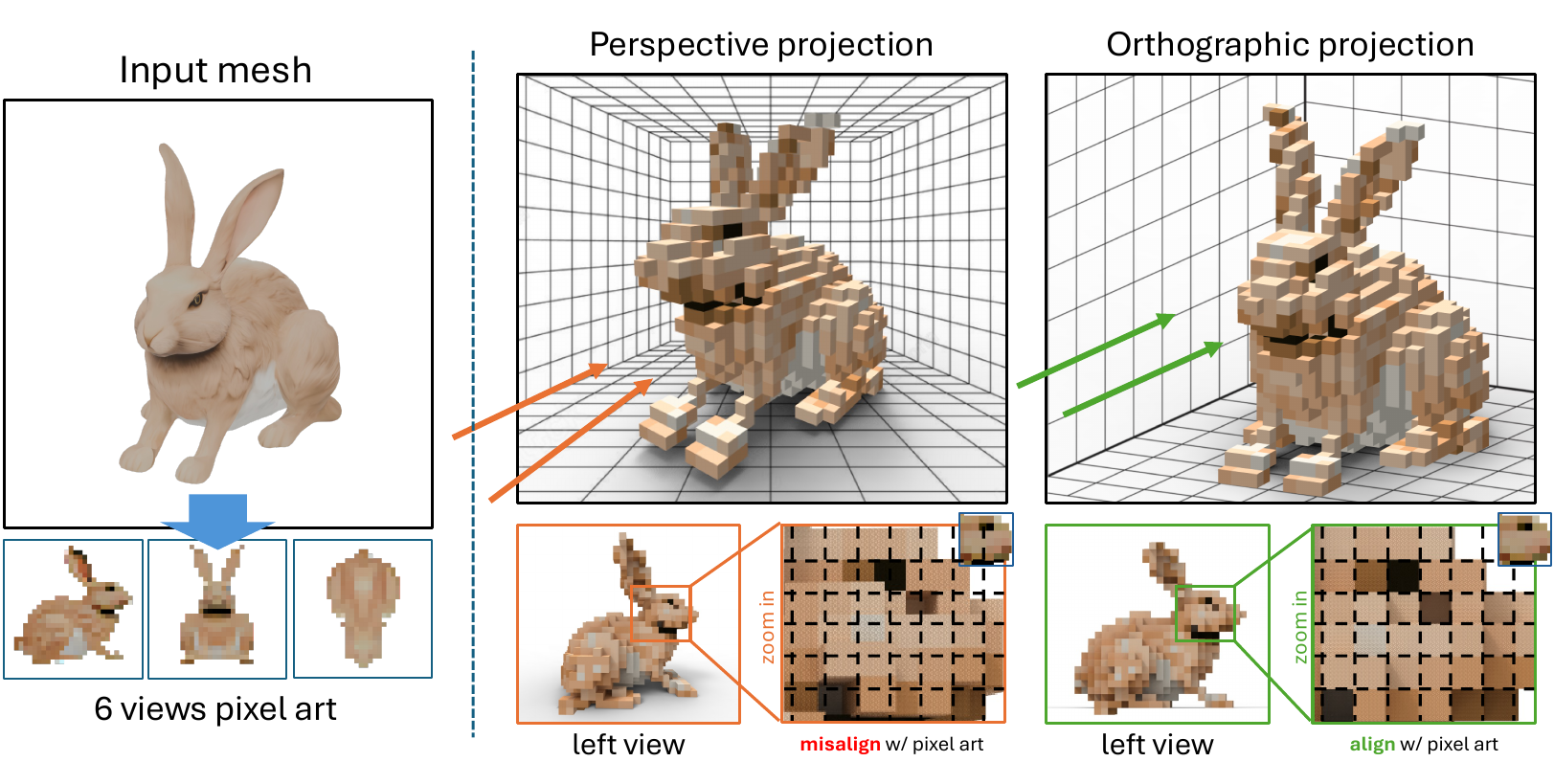}
\vspace{-6mm}
\caption{\textbf{Perspective vs. Orthographic.} 
(\emph{Left}) Six-view pixel art pipeline. (\emph{Right}) Perspective views (\textcolor{red}{red}) misalign pixels, while six orthographic views (\textcolor{teal}{green}) enable precise pixel–voxel alignment.}
\label{fig:orthographic}
\end{figure}

To utilize the abstract features and clean edges of pixel art for 3D grid supervision, we fine-tune the voxel space by rendering orthographic projections from six axis-aligned views and comparing them against pixel art supervision generated by the pixel art generator~\citep{wu2022make}. This six-view setup compactly covers the major surfaces of the object, while orthographic rendering formulates parallel ray casting $\mathbf{r}_i(t) = \mathbf{o}_i + t\mathbf{d}$, where $\mathbf{o}_i$ is the ray origin of pixel $\mathbf{p}_i$ and $\mathbf{d}$ is the fixed ray direction. All rays are parallel, ensuring pixel-to-voxel alignment without perspective distortions (Fig.~\ref{fig:orthographic}).

We apply two foundational losses to supervise geometry and structure:
\begin{equation}
\mathcal{L}_{\text{pixel}} = \left\|C(\mathbf{r}) - C_{\text{pixel}}\right\|_2^2
\end{equation}
\begin{equation}
\mathcal{L}_{\text{depth}} = \left\|D(\mathbf{r}) - D_{\text{gt}}\right\|_1,
\end{equation}
where $C(\mathbf{r})$ and $D(\mathbf{r})$ are the rendered color and depth along ray $\mathbf{r}$, 
$C_{\text{pixel}}$ is the RGB color from the pixel-art supervision, 
and $D_{\text{gt}}$ is the mesh-projected depth.

We also use an alpha loss to suppress density in background regions, enforcing background transparency to avoid floating density artifacts:
\begin{equation}
\mathcal{L}_{\alpha} = \left\|\mathcal{M}_\alpha \odot \bar{\alpha}\right\|^2,
\end{equation}
where $\mathcal{M}_\alpha \in \{0,1\}^{H\times W}$ is a binary mask from the pixel art alpha channel (1 for background), 
and $\bar{\alpha}$ denotes the accumulated ray opacity from volume rendering, which is encouraged to be $0$ for background rays to allow full transparency. 
This encourages transparent regions in the pixel art to remain fully transmissive, preventing the formation of undesired voxels in areas without valid supervision.

By leveraging pixel art as the supervision signal, each voxel grid more effectively captures and expresses the most important structural and appearance information.


\subsection{CLIP-based Semantic Loss}
\label{sec:clip} 
To incorporate semantic supervision, we sample half of the total rays to form patches for computing a CLIP-based perceptual loss. 
During training, we randomly sample patch rays \( (\mathbf{o}_{\text{patch}}, \mathbf{d}_{\text{patch}}) \) from rendered images \( I_{\text{mesh}} \) of input mesh. 
Given the rendered patch \( \hat{I}_{\text{patch}} \) and the corresponding mesh-based patch \( I^{\text{mesh}}_{\text{patch}} \), we extract CLIP features~\citep{radford2021learning, frans2022clipdraw} and compute a perceptual loss via cosine similarity:
\begin{equation}
\mathcal{L}_\text{clip} = 1 - \cos\left( \text{CLIP}(\hat{I}_{\text{patch}}), \ \text{CLIP}(I^{\text{mesh}}_{\text{patch}}) \right),
\end{equation}
where cosine similarity is defined as 
\(\cos(a,b) = \frac{\langle a, b \rangle}{\|a\| \, \|b\|}\), 
and $CLIP(\cdot)$ denotes the CLIP image encoder output.
This loss encourages voxel-rendered outputs to remain semantically aligned with the input mesh while supporting stylized abstraction, as illustrated in stage (c) of Fig.~\ref{fig:pipeline}.

\subsection{Discrete Color Selection via Gumbel-Softmax}
\label{sec:gumbel} 
To generate clean and stylized voxel appearances while allowing flexible color selection strategies, we adopt a palette-based quantization scheme where each voxel selects a color from a predefined palette. This palette is extracted from the six-view pixel art images using a chosen clustering method before Gumbel-Softmax quantization. 

Instead of regressing RGB values, each voxel \( (i, j, k) \) stores a color logit vector \( \boldsymbol{\lambda}_{i,j,k} \in \mathbb{R}^C \), where \( C \) is the number of discrete colors in the predefined palette.

During training, Gumbel noise \( \mathbf{G}_{i,j,k} \sim \text{Gumbel}(0,1) \in \mathbb{R}^C \) is added to produce noisy logits:
\begin{equation}
\mathbf{Y}_{i,j,k} = \boldsymbol{\lambda}_{i,j,k} + \mathbf{G}_{i,j,k},
\end{equation}
where \(Y_{i,j,k,n}\) denotes the noisy logit for the \(n\)-th palette color at voxel \((i,j,k)\), with \(n \in \{1,\dots,C\}\). 
A temperature-controlled softmax~\citep{jang2017categorical,maddison2017concrete} is then applied:
\begin{equation}
s_{i,j,k,n}(\tau) = \frac{\exp(Y_{i,j,k,n} / \tau)}{\sum_{n'=1}^{C} \exp(Y_{i,j,k,n'} / \tau)},
\end{equation}
where \(s_{i,j,k,n}(\tau)\) is the probability of selecting the \(n\)-th color in the palette for voxel \((i,j,k)\), and \(\tau\) is the temperature parameter controlling distribution sharpness.

In early training, we use the soft distribution \( s_{i,j,k} \) directly. 
Later, we switch to the straight-through variant, where the forward pass uses a one-hot selection at \( \arg\max_n s_{i,j,k} \), while gradients are backpropagated through the soft weights. 
We anneal the temperature \( \tau \) during training to encourage smooth exploration in the early stages and sharper, more discrete selections later.  
The sampled RGB value is computed as:
\begin{equation}
\text{RGB}_{i,j,k} = \sum_{n=1}^{C} s_{i,j,k,n} \cdot \mathbf{c}_n,
\end{equation}
where \( \mathbf{c}_n \in \mathbb{R}^3 \) is the \(n\)-th color in the palette.
 
After training, we directly select the color with the highest logit:
\begin{equation}
\text{RGB}_{i,j,k}^{\text{voxel}} = \mathbf{c}_{\arg\max\limits_{n} \ \lambda_{i,j,k,n}},
\end{equation}
producing fully discrete voxel outputs. 
This process is illustrated in stage~(d) of Fig.~\ref{fig:pipeline}.

To enhance flexibility in stylization, this strategy allows users to choose the color selection method and number of colors, enabling explicit control over both color richness and overall style of the voxel art, making the design process more aligned with practical usage scenarios.

\subsection{Loss Summary and Training Procedure}
The overall loss optimized during fine-tuning is a weighted sum of multiple components that jointly supervise pixel-art faithfulness, geometry consistency, semantic alignment, and spatial regularity:
\begin{equation}
\mathcal{L}_\text{total} = 
\lambda_\text{pixel} \cdot \mathcal{L}_\text{pixel} +
\lambda_\text{depth} \cdot \mathcal{L}_\text{depth} +
\lambda_\text{alpha} \cdot \mathcal{L}_\text{alpha} +
\lambda_\text{clip} \cdot \mathcal{L}_\text{clip},
\end{equation}
where $\mathcal{L}_\text{pixel}$, $\mathcal{L}_\text{depth}$, and $\mathcal{L}_\text{clip}$ encourage pixel-level accuracy, depth consistency, and semantic alignment, respectively, while $\mathcal{L}_\text{alpha}$ suppresses background opacity to yield clean silhouettes. In Stage~2, rays are split into two groups: (1) $\mathcal{L}_\text{pixel}$, $\mathcal{L}_\text{depth}$, and $\mathcal{L}_\text{alpha}$, and (2) $\mathcal{L}_\text{clip}$ on rendered patches, all computed via volumetric rendering (\cref{eq:vol_render}). Thus, geometric supervision of the density grid is provided by $\mathcal{L}_\text{pixel}$, $\mathcal{L}_\text{depth}$, and $\mathcal{L}_\text{alpha}$, while semantic supervision comes from $\mathcal{L}_\text{clip}$, which guides voxel appearance toward the intended pixel-art style.

\section{Experiments}
\subsection{Experimental Setup}
\paragraph{Dataset.}  
We evaluate our method on three mesh datasets:
\textbf{Rodin} \citep{wang2022rodin}, 
\textbf{Unique3D}~\citep{wu2024unique3d}, and \textbf{TRELLIS}~\citep{xiang2024structured}. 
Rodin and Unique3D primarily feature character 3D assets with rich semantic details, making them ideal for evaluating voxel abstraction and stylized representation. We also evaluate on diverse categories including architecture and vehicles; see supplementary material for details.

\paragraph{Implementation details.}  
Training follows a two-stage schedule: \textbf{(a) Coarse Voxelization}: optimize the voxel grid for 8000 iterations to capture global structure;  
\textbf{(b) Pixel Art Supervision}: fine-tune for 6500 iterations with MSE, Depth, and CLIP losses on six orthographic views, rendered at a resolution of $1200\times1200$, using fixed $80\times80$ patches randomly sampled each iteration for CLIP loss.  
In the final 2000 iterations, supervision is applied only to the front view to enhance key abstract features.  
Gumbel-Softmax sampling is performed over a fixed palette, with temperature $\tau$ annealed from 1.0 to 0.1.


\paragraph{Baseline methods.}  
We compare against:  
\begin{enumerate}
    \item \textbf{Pixel art to 3D extension}: Render the input mesh into images, stylize them into pixel art, then train the original DVGO with these pixel-art images, using the coarse voxel grid as the final output.
    \item \textbf{IN2N}~\citep{haque2023instruct}: Language-guided mesh editing with view-consistent 3D stylization.  
    \item \textbf{Vox-E}~\citep{sella2023vox}: Language-to-voxel generation prioritizing semantics over fine geometry.  
    \item \textbf{Blender Geometry Nodes}: Procedural mesh-to-voxel conversion, fast but without semantic or stylization control.  

\end{enumerate}


\subsection{Qualitative Comparisons}

We qualitatively compare our method with Pixel art to 3D extension, IN2N, Vox-E, and Blender on eight character meshes from the evaluation datasets (Fig.~\ref{fig:comparison}), with an additional eight groups of comparisons provided in the supplementary material. 

IN2N preserves coarse structure but suffers from large variations across different guidance images, often failing to produce consistent voxelized results; Vox-E yields smoother volumes yet misses the discrete, blocky style of voxel art; Blender produces clean abstraction through procedural voxelization, which is akin to simple downsampling, but requires manual tuning and lacks semantic alignment.

Our method preserves key cues (e.g., ears, eyes) with sharp edges across 25$\times$–50$\times$ resolutions, achieving both expressive stylization and semantic fidelity. Additional results are provided in the supplementary material.

\begin{figure*}[t]
  \centering
  \includegraphics[width=\textwidth]{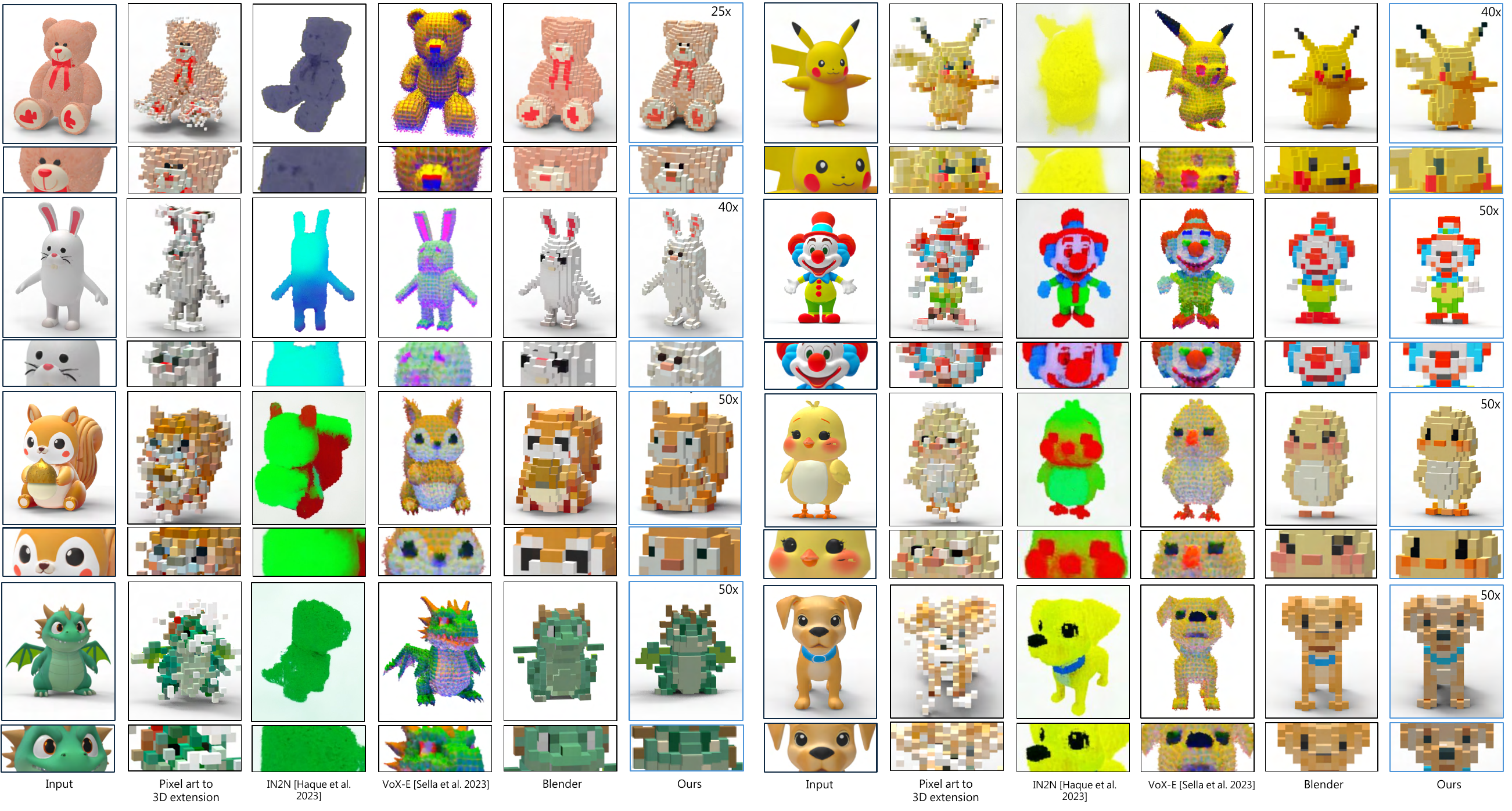}
\vspace{-3mm}
  \caption{\textbf{Qualitative comparisons on character models from the Rodin~\citep{wang2022rodin} dataset.} We compare our voxel art results with Pixel art to 3D extension, IN2N~\citep{haque2023instruct}, Vox-E~\citep{sella2023vox}, and Blender's voxelization. Our method produces stylized yet consistent voxel representations with pixel art aesthetics.
  }
  \label{fig:comparison}
\end{figure*}

\subsection{Quantitative Comparisons}

\begin{table}[t]
\small
  \centering
  \setlength{\tabcolsep}{6pt}
  \renewcommand{\arraystretch}{1.1}
  \caption{\textbf{Average CLIP-IQA scores over all 35 examples.} 
  Best scores are \textbf{highlighted}.}
  \label{tab:clip-iqa-35}
  \begin{tabular}{lccccc}
  \toprule
  Method & Pixel & IN2N & Vox-E & Blender & Ours \\
  \midrule
  CLIP-IQA & 35.53 & 23.93 & 35.02 & 36.31 & \textbf{37.12} \\
  \bottomrule
  \end{tabular}
\end{table}

To assess stylization fidelity and semantic preservation, we adopt the CLIP-IQA framework. 
For each character, we use GPT-4 to generate a detailed textual description based on the original mesh images, 
prepended with ``A voxel art of...'' (e.g., ``A voxel art of a pink teddy bear with a red bow and heart-shaped feet''). 
We use OpenAI’s ViT-B/32 CLIP model and compute the average cosine similarity between each prompt and the rendered images from different methods.


As shown in Table~\ref{tab:clip-iqa-35}, the reported CLIP-IQA scores are averaged over all \textbf{35} cases. Our method consistently achieves the highest score, demonstrating superior semantic alignment and stylized abstraction across a diverse set of character meshes.


\begin{figure*}[t]
\centering
\includegraphics[width=1.0\textwidth]{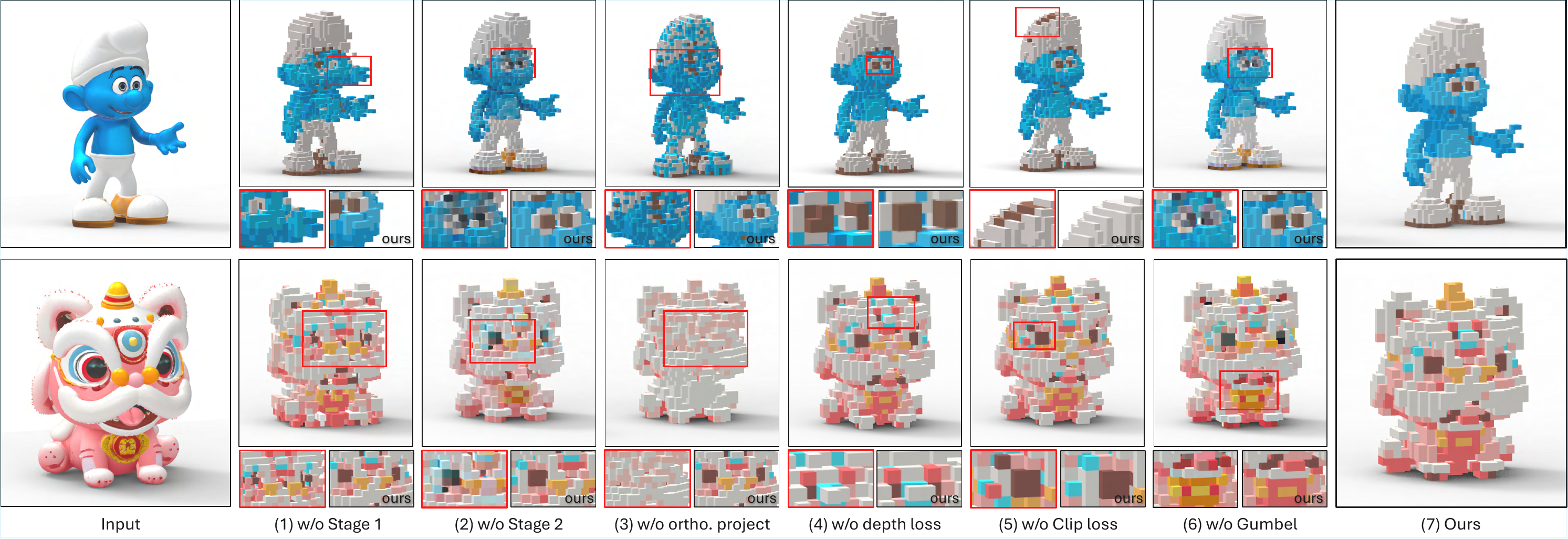}
\vspace{-6mm}
\caption{\textbf{Ablation study on key components.} 
We evaluate the impact of each module by removing it individually: 
(1) w/o Stage 1, 
(2) w/o Stage 2, 
(3) w/o orthographic projection (replaced with perspective projection), 
(4) w/o depth loss, 
(5) w/o CLIP loss, and 
(6) w/o Gumbel Softmax (resulting in continuous and unconstrained colors). 
Each row shows a different input, and zoom-in regions highlight local differences. 
Removing each component leads to noticeable degradation in geometry, color consistency, or semantic fidelity, while the full model produces coherent and stable voxel stylization.}
\label{fig:ablation}
\end{figure*}

\begin{figure}[t]
\centering
\includegraphics[width=\columnwidth]{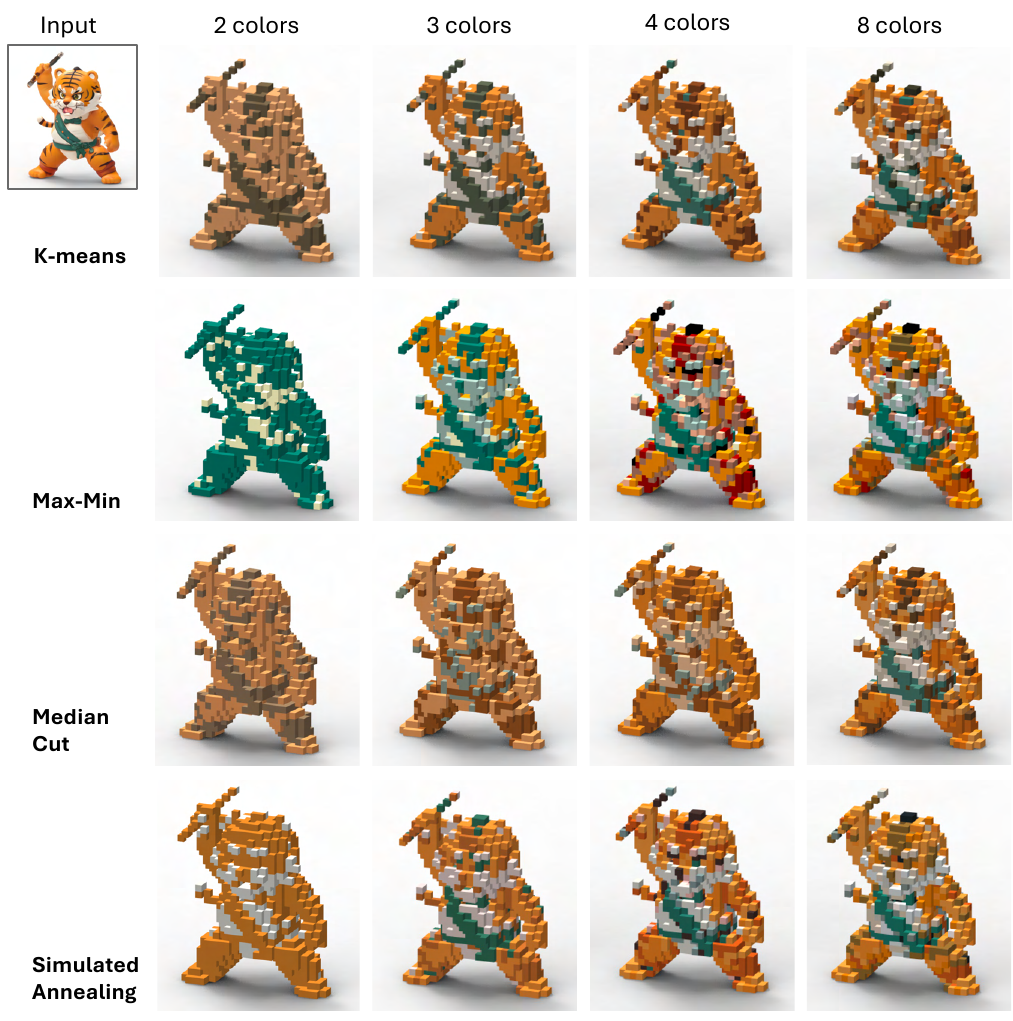}
\vspace{-2mm}
\caption{\textbf{Effect of Palette Selection and Color Count.}
Each row corresponds to a different palette extraction method: K-means, Max-Min, Median Cut, and Simulated Annealing. Each column shows increasing color counts (2, 3, 4, 8). Each method produces unique color clustering effects.}
\label{fig:color_palette}
\vspace{-3mm}
\end{figure}

\subsection{Ablation Study}

\begin{table}[t]
\vspace{-2mm}
\centering
\small
\caption{\textbf{Ablation study of model components (CLIP-IQA).} 
We evaluate the impact of each component by removing it individually. 
Our full model achieves the best performance. 
Removing stage design or orthographic projection leads to significant degradation, 
while depth loss, CLIP loss, and Gumbel Softmax contribute to consistent improvements in semantic alignment.}
\label{tab:ablation}
\resizebox{1.0\columnwidth}{!}{
\begin{tabular}{l|cccccc|c}
\toprule
Method & Stage 1 & Stage 2 & Ortho. Proj. & Depth loss & CLIP loss & Gumbel & CLIP-IQA \\
\midrule
(1) w/o Stage 1 &  & \checkmark & \checkmark & \checkmark & \checkmark & \checkmark & 28.42 \\
(2) w/o Stage 2 & \checkmark &  &  &  &  &  & 34.32 \\
(3) w/o ortho. proj. & \checkmark & \checkmark &  & \checkmark & \checkmark & \checkmark & 27.38 \\
(4) w/o depth loss & \checkmark & \checkmark & \checkmark &  & \checkmark & \checkmark & 39.75 \\
(5) w/o CLIP loss & \checkmark & \checkmark & \checkmark & \checkmark &  & \checkmark & 39.23 \\
(6) w/o Gumbel & \checkmark & \checkmark & \checkmark & \checkmark & \checkmark &  & 39.31 \\
\rowcolor{black!10}
(7) Ours & \checkmark & \checkmark & \checkmark & \checkmark & \checkmark & \checkmark & \textbf{40.06} \\
\bottomrule
\end{tabular}
}
\vspace{-3mm}
\end{table}

We analyze the contribution of each component by removing one module at a time. Qualitative results are shown in Fig.~\ref{fig:ablation}, and quantitative results are reported in Table~\ref{tab:ablation}.

\paragraph{Stage-wise optimization.}
Without Stage 1, the model lacks proper geometric initialization, leading to distorted shapes. Without Stage 2, the result degenerates to a coarse DVGO grid without abstraction or semantic refinement. This shows that Stage 1 provides geometric stability, while Stage 2 is the key component that enables semantic abstraction and stylized voxel representation.

\paragraph{Orthographic projection.}
Replacing orthographic projection with perspective projection causes severe color misalignment, as pixel colors no longer correspond to voxel locations, resulting in inconsistent appearance (Fig.~\ref{fig:ablation}).

\paragraph{Depth loss.}
Without depth loss, the model produces plausible views individually but fails to preserve global 3D structure, leading to geometric distortions.

\paragraph{CLIP loss.}
Removing CLIP loss reduces semantic clarity and leads to less coherent color regions, highlighting the importance of semantic alignment.

\paragraph{Gumbel-Softmax.}
Without Gumbel-Softmax, colors become mixed and lack clear boundaries, failing to produce clean and discrete voxel-style color patterns.

\paragraph{Summary.}
Fig.~\ref{fig:ablation} and Table~\ref{tab:ablation} show that all components contribute to geometry, semantic alignment, and discrete color representation. Our full model achieves the best performance (40.06 CLIP-IQA). Removing any component consistently degrades quality. The quantitative results in Table~\ref{tab:ablation} are averaged over five objects.

\paragraph{Palette controllability.}
We analyze palette design by varying the number of colors (2, 3, 4, 8) and extraction methods (K-means, K-means with rare color boosting, Median Cut, Max-Min, and Simulated Annealing). 

The palette is extracted from all pixels across the six input pixel-art views to form a shared discrete color set for voxel optimization.

As shown in Fig.~\ref{fig:color_palette}, increasing palette size improves color richness, while different strategies affect color distribution. K-means favors dominant tones, while the rare-color variant preserves infrequent details, and other methods (Median Cut, Max-Min, Simulated Annealing) promote balanced or diverse palettes.

Smaller palettes lead to stronger abstraction, whereas larger palettes improve detail. Our method maintains consistent voxel structure while enabling flexible control over color abstraction. Additional results and detailed palette selection strategies are provided in the supplementary material.

\begin{figure*}[t]
\centering
\includegraphics[width=\textwidth]{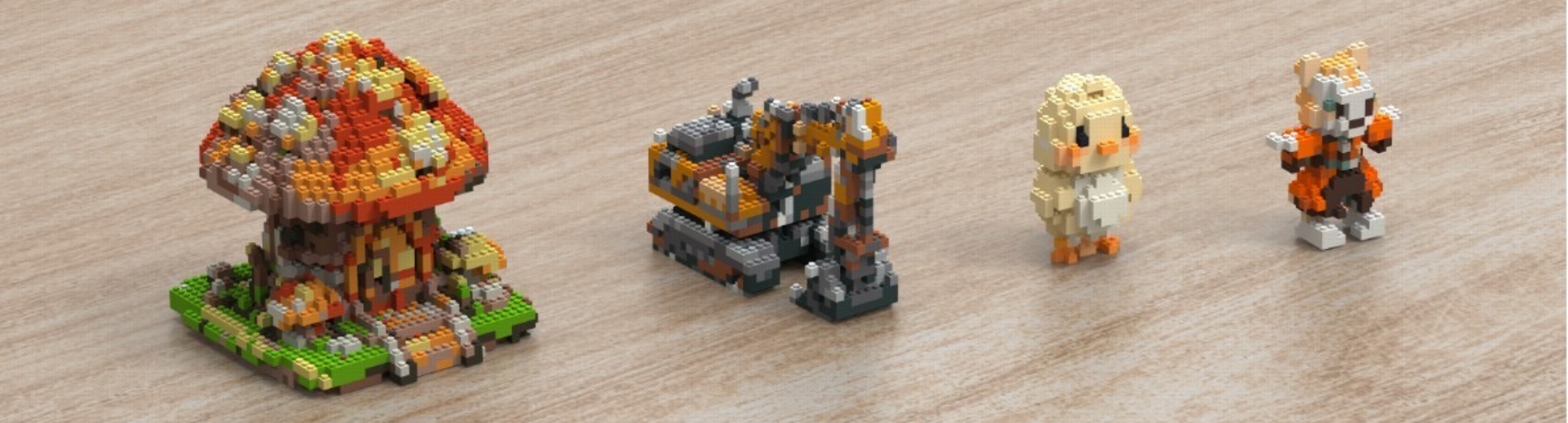}
\caption{\textbf{Fabrication: LEGO render.} Rendered using KeyShot 2023. Our method extends to LEGO applications, where achieving rich visual results within the limited color palette is crucial for practical fabrication.}
\label{fig:lego_render}
\end{figure*}

\begin{table}[t]
  \centering
  \small
  \captionsetup{type=table}
  \caption{\textbf{User studies.} 
  (a) 35 examples (72 participants). (b) Color quantization (10 art-trained).}
  \label{tab:wrap-user}
  \begin{minipage}[t]{0.52\linewidth}
    \centering
    \subcaption*{\small (a) Image quality (user votes, \%)}
\resizebox{\textwidth}{!}{
    \begin{tabular}{@{}lccc@{}}
      \toprule
      Metric & Abstract & Appeal & Geometry \\
      \midrule
      Ours   & \textbf{77.90} & \textbf{80.36} & \textbf{96.55} \\
      Others & 22.10 & 19.64 & 3.45 \\
      \bottomrule
    \end{tabular}
    }
  \end{minipage}
  \hfill 
  \begin{minipage}[t]{0.45\linewidth}
    \centering
    \subcaption*{\small (b) Color quantization preference (\%)}
\resizebox{\textwidth}{!}{
    \begin{tabular}{@{}lcc@{}}
      \toprule
       & w/o Gumbel & w/ Gumbel \\
      \midrule
      Preferred & 11.11 & \textbf{88.89} \\
      \bottomrule
    \end{tabular}
    }
  \end{minipage}
\end{table}

\subsection{User Study}

We conducted a user study with 72 participants to evaluate our method against four baselines: Pixel Art to 3D extension, IN2N~\citep{haque2023instruct}, Vox-E~\citep{sella2023vox}, and Blender Geometry Nodes.

The study consists of two parts. In the first part (35 examples), participants evaluated \emph{abstract detail} and \emph{visual appeal}. In the second part (geometry evaluation), participants compared grayscale renderings to assess shape preservation. 

As shown in Table~\ref{tab:wrap-user} (a), our method is consistently preferred, achieving 77.90\%, 80.36\%, and 96.55\% in abstractness, appeal, and geometry, respectively.

\paragraph{Expert Study on Color Preference.}
We further conduct a focused evaluation on color quantization with 10 art-trained participants. As shown in Table~\ref{tab:wrap-user} (b), 88.89\% of participants prefer the results with Gumbel-Softmax, demonstrating its effectiveness in producing clearer structures and more distinct color regions.

More details on study design, questions, and additional results are provided in the supplementary material.


\section{Conclusion}
We introduce \textbf{Voxify3D}, a novel framework for transforming 3D meshes into stylized voxel art with strong semantic abstraction and structural consistency. By combining coarse voxel optimization, orthographic pixel art supervision, and palette-based color quantization, our method achieves expressive and visually appealing results across a variety of character assets. Extensive experiments and user studies confirm its advantages over existing baselines in both geometric faithfulness and artistic stylization. 

In addition to digital results, we demonstrate the fabrication potential of our voxel outputs via LEGO-style assemblies (Fig.~\ref{fig:lego_render}), enabled by their discrete structure and limited color palette.

\paragraph{Limitations and Future Work.}
Voxify3D struggles with highly intricate shapes, where thin structures or fine facial details may be lost at low voxel resolutions. Future work may explore integrating geometric priors or training strategies to enhance detail preservation and scalability, as well as adopting assembly-aware fabrication strategies inspired by LEGO brick design and connection principles to improve the physical realizability of voxel-based models.

\newpage
\paragraph{Acknowledgements.}
This research was funded by the National Science and Technology Council, Taiwan, under Grants NSTC 112-2222-E-A49-004-MY2 and 113-2628-E-A49-023-. The authors are grateful to Google, NVIDIA, and MediaTek Inc. for their generous donations. Yu-Lun Liu acknowledges the Yushan Young Fellow Program by the MOE in Taiwan.

{\small
\bibliographystyle{ieeenat_fullname}
\bibliography{11_references}
}

\ifarxiv \clearpage \appendix \section{Overview}
\label{sec:Overview}
This supplementary material provides additional details that complement our main paper. We include implementation details (Sec.~\ref{sec:details}) covering our codebase and training architecture, pixel art generator, logit grid initialization, parameter settings, loss design, temperature annealing schedule, cross-view inconsistency handling, and palette selection strategies. We also provide experimental information (Sec.~\ref{sec:information}) including our CLIP-IQA evaluation protocol, user study details, expert study on color preference, and run time analysis. Additionally, we present additional qualitative results (Sec.~\ref{sec:additional_results}) with more comparisons against baselines, results with varying palette settings, and results under different voxel sizes. We also provide comparisons with recent voxel art generation methods, including Gemini 3~\cite{google_gemini} and Rodin~\cite{wang2022rodin}. Finally, we show failure cases and analyze potential future directions (Sec.~\ref{sec:failure_cases}).

\section{Implementation Details}
\label{sec:details}

\vspace{3pt}
\noindent {\bf Codebase and training architecture.} 
Our implementation builds on DVGO~\cite{sun2022direct}. 
We adopt a two-stage training pipeline. 
In \textit{Stage~1}, we follow DVGO to train a coarse voxel grid, which initializes both color and density representations. 
In \textit{Stage~2}, the input consists of six orthographic views stylized into pixel art. 
Using orthographic projection, each pixel from the pixel art is directly aligned with the voxel grid, ensuring per-pixel to voxel correspondence. 
After 4500 iterations, training is restricted to the front view, which typically contains the most salient semantic features (e.g., facial structures), allowing the model to refine key abstract details while maintaining consistency from the earlier multi-view supervision.

\vspace{3pt}

\noindent {\bf Pixel art generator.}

Our pipeline requires stylized pixel art inputs rather than simple low-resolution downsampling.  

We adopt the MYOS~\cite{wu2022make} generator to transform mesh renderings into high-quality pixel art, 
which preserves sharp boundaries and stylized abstractions.  
As illustrated in Fig.~\ref{fig:pixelart}, 
naïve downsampling produces blurry textures, while MYOS yields pixelated structures with clear edges, 
better aligned with voxel abstraction.

\begin{figure}[t]
    \centering
    \includegraphics[width=\linewidth]{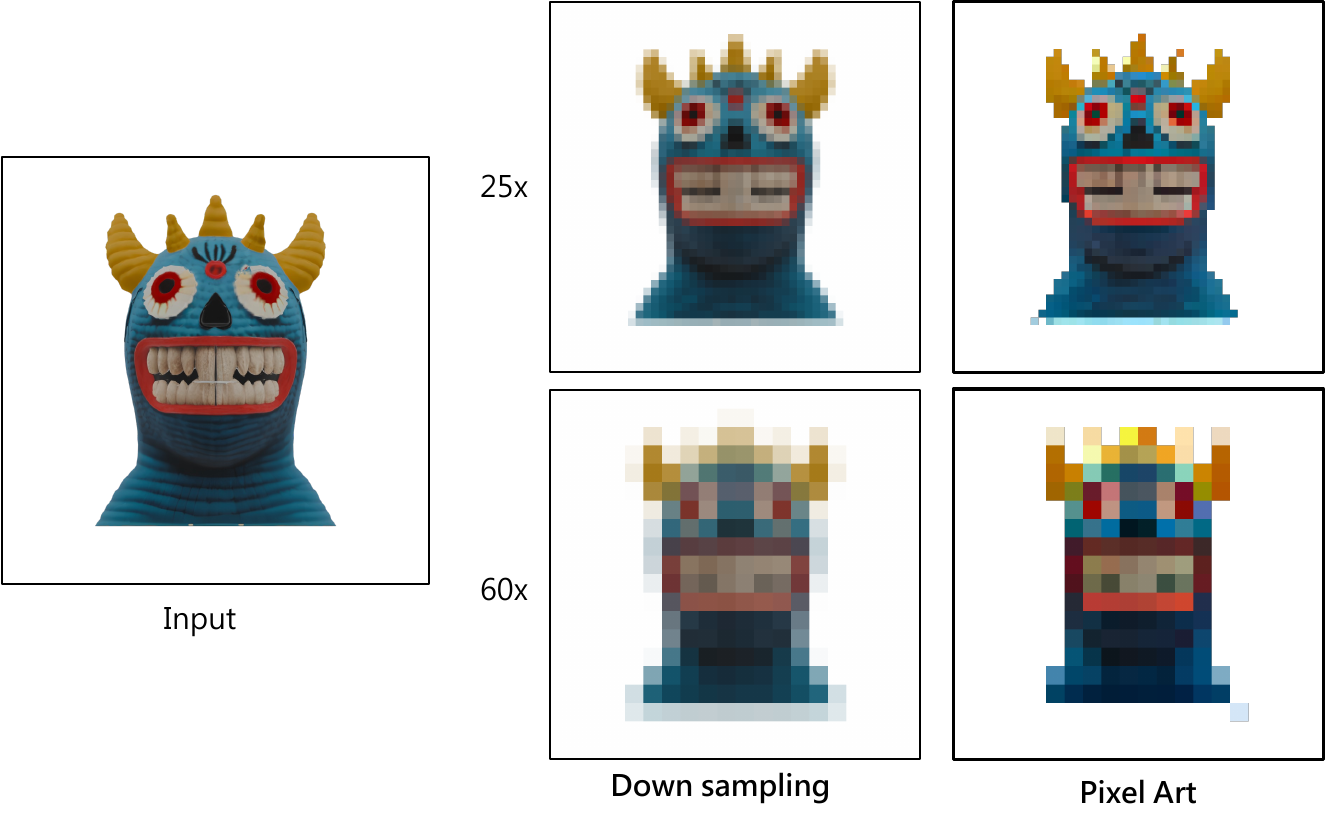}
    \caption{\small \textbf{Downsample vs. Pixel Art}}
    \label{fig:pixelart}
\end{figure}

\vspace{3pt}
\noindent {\bf Logit grid initialization.} 
In \textit{Stage~2}, we initialize each voxel’s logit vector by the negative distance between its \textit{Stage~1} RGB color and the palette entries. 
This provides a stable bias toward closer colors and converges better than random initialization.

\vspace{3pt}
\noindent {\bf Parameter settings.} 
We summarize the key training parameters for Stage~1 (voxel grid initialization) and Stage~2 (logit grid optimization). 
\begin{table*}[t]
\centering
\caption{\textbf{Training parameters for Stage~1 (left) and Stage~2 (right).}}
\label{tab:train-params}
\vspace{-3mm}
\begin{minipage}[t]{0.49\linewidth}
\centering
\setlength{\tabcolsep}{6pt}
\renewcommand{\arraystretch}{1.15}
\begin{tabular}{lc}
\toprule
Parameter & Value \\
\midrule
Iterations ($N_{\text{iters}}$) & 8000 \\
Batch size ($N_{\text{rand}}$) & 8192 \\
Learning rate (density grid) & $1\times 10^{-1}$ \\
Learning rate (color grid $k_0$) & $1\times 10^{-1}$ \\
LR decay step & 20 \\
\bottomrule
\end{tabular}
\end{minipage}\hfill
\begin{minipage}[t]{0.49\linewidth}
\centering
\setlength{\tabcolsep}{6pt}
\renewcommand{\arraystretch}{1.15}
\begin{tabular}{lc}
\toprule
Parameter & Value \\
\midrule
Iterations ($N_{\text{iters}}$) & 6500 \\
Batch size ($N_{\text{rand}}$) & 8192 \\
Learning rate (density grid) & $5\times 10^{-3}$ \\
Learning rate (logit grid) & $1\times 10^{-1}$ \\
LR decay step & 20 \\
\bottomrule
\end{tabular}
\end{minipage}
\vspace{0mm}
\end{table*}

\noindent {\bf Loss design.} 
We adopt different objectives across the two training stages. 

\textit{Stage~1 (Coarse voxelization).}  
The voxel grid is optimized with MSE reconstruction loss, regularized by density and background terms:
\[
\mathcal{L}_\text{total} = 
\mathcal{L}_\text{render} +
\lambda_d \mathcal{L}_\text{density} +
\lambda_b \mathcal{L}_\text{bg},
\]
where $\mathcal{L}_\text{render}$ is MSE between rendered and target colors, 
$\mathcal{L}_\text{density}$ applies density regularization and total variation smoothing, 
and $\mathcal{L}_\text{bg}$ uses entropy to suppress background noise.  
This stage provides a stable initialization for both color and density.

\textit{Stage~2 (Pixel-art supervision).}  
The fine-tuning objective combines pixel accuracy, geometry regularization, semantic alignment, and silhouette clarity:
\[
\mathcal{L}_\text{total} =
\lambda_\text{pixel} \mathcal{L}_\text{pixel} +
\lambda_\text{depth} \mathcal{L}_\text{depth} +
\lambda_\text{alpha} \mathcal{L}_\text{alpha} +
\lambda_\text{clip} \mathcal{L}_\text{clip}.
\]

\noindent {\bf Implementation details.} 
$\mathcal{L}_\text{pixel}$ (MSE) is up-weighted to ensure faithful color abstraction.  
$\mathcal{L}_\text{depth}$ is scaled by voxel resolution: $20$ normally, and increased to $30$ after step 4500.  
$\mathcal{L}_\text{alpha}$ enforces clean silhouettes via transparency regularization.  
$\mathcal{L}_\text{clip}$ is applied until step 6000, using $80\times80$ patches per iteration for semantic alignment.  
After step 6000, optimization focuses mainly on background transparency ($\mathcal{L}_\text{alpha}$), while CLIP loss is disabled. This scheduling ensures early semantic guidance, followed by refinement of geometry and silhouettes. The detailed training parameters are included in Tab.~\ref{tab:train-params}, and the specific loss weights are detailed in Tab.~\ref{tab:loss-weights}.

\begin{table*}[t]
\centering
\renewcommand{\arraystretch}{1.2}
\caption{\textbf{Loss weights used in our implementation.}}
\label{tab:loss-weights}
\begin{tabular}{cccccc}
\toprule
$\lambda_\text{pixel}$ & $\lambda_\text{depth}$ & $\lambda_\text{alpha}$ & $\lambda_\text{clip}$ & $\lambda_b$ & $\lambda_d$ \\
\midrule
$\times 10$ & \makecell{10 / 20 \\ ($30$ after 4500 iter)} & $\times 20$ & \makecell{$\times 1$ \\ (until 6000 iter)} & \makecell{$\times 0.5$ \\ (Stage~1)} & \makecell{$\times 0$ default \\ (Stage~1)} \\
\bottomrule
\end{tabular}
\end{table*}

\vspace{3pt}
\noindent {\bf Temperature annealing schedule.}  
We apply a step-wise annealing schedule for the Gumbel-Softmax 
temperature $\tau$, gradually lowering it to encourage sharper palette selection as training progresses. 
The temperature starts high to allow exploration of multiple colors, and progressively decreases to enforce deterministic palette assignments toward convergence.
The complete annealing schedule is shown in Tab.~\ref{tab:tau-schedule}.
\begin{table*}[t]
\centering
\caption{\textbf{Step-wise annealing schedule of the Gumbel-Softmax temperature $\tau$ during Stage~2.}}
\label{tab:tau-schedule}
\begin{tabular}{cccccc}
\toprule
$<1000$ & $1000$–$2999$ & $3000$–$3999$ & $4000$–$4999$ & $5000$–$6000$ & $>6001$ \\
\midrule
1.0 & 0.8 & 0.3 & 0.6 & 0.3 & 0.1 \\
\bottomrule
\end{tabular}
\end{table*}

\vspace{3pt}
\noindent {\bf Cross-view inconsistency.}  
Supervision from six orthographic views keeps inconsistencies minimal, mostly near boundaries. 
To further refine salient cues, the last 2000 iterations are trained only on the front view (rich in facial details), reinforcing key features while preserving global consistency from earlier multi-view supervision.

\vspace{3pt}
\noindent {\bf Palette selection strategies.}  
We explored multiple strategies for extracting compact color palettes from input images:  

\begin{itemize}[leftmargin=12pt,itemsep=0pt,topsep=2pt]
    \item \textit{K-means clustering}: baseline method that partitions pixels into $C$ clusters and uses centroids as representative colors.  
    \item \textit{K-means with rare color boosting}: explicitly incorporates infrequent colors to prevent palette collapse into dominant tones.  
    \item \textit{Median cut}: recursively splits the RGB space by channel ranges to ensure balanced coverage of color distributions.  
    \item \textit{Max–min picking}: iteratively selects farthest colors in feature space to maximize palette diversity.  
    \item \textit{Simulated annealing}: formulates palette extraction as a discrete optimization problem, refining palettes via stochastic search.  
\end{itemize}

\section{Experimental Information}
\label{sec:information}

\noindent {\bf CLIP-IQA evaluation protocol. } 
We evaluate stylization fidelity and semantic preservation using CLIP-IQA. 
Text prompts (``A voxel art of...'') are generated by GPT-4 from input mesh images, and cosine similarity is computed between the prompts and rendered results using CLIP (ViT-B/32), averaged over 35 cases. 
While training employs CLIP loss in an image--image setting, evaluation is conducted with text prompts, ensuring that CLIP-IQA reflects semantic fidelity rather than overfitting to the training objective. 

Furthermore, we report CLIP-IQA scores across multiple voxel resolutions (Tab.~\ref{tab:clip-ablation}), demonstrating that our method consistently maintains semantic alignment under different discretization levels. 

\begin{table}[t]
  \centering
  \small
  \caption{\textbf{CLIP-IQA ablation across voxel sizes.} 
  CLIP loss improves semantic alignment consistently.}
  \label{tab:clip-ablation}
  \begin{tabular}{lcccc}
  \toprule
  Voxel Size & 25$\times$ & 30$\times$ & 40$\times$ & 50$\times$ \\
  \midrule
  w/o CLIP Loss & 40.89 & 40.55 & 38.92 & 38.64 \\
  \rowcolor{black!10}
  w/ CLIP (ours) & \textbf{41.35} & \textbf{41.03} & \textbf{40.07} & \textbf{40.14} \\
  \bottomrule
  \end{tabular}
\end{table}

\vspace{3pt}
\noindent {\bf User study details.}

\begin{figure}
    \centering
    \includegraphics[width=\linewidth]{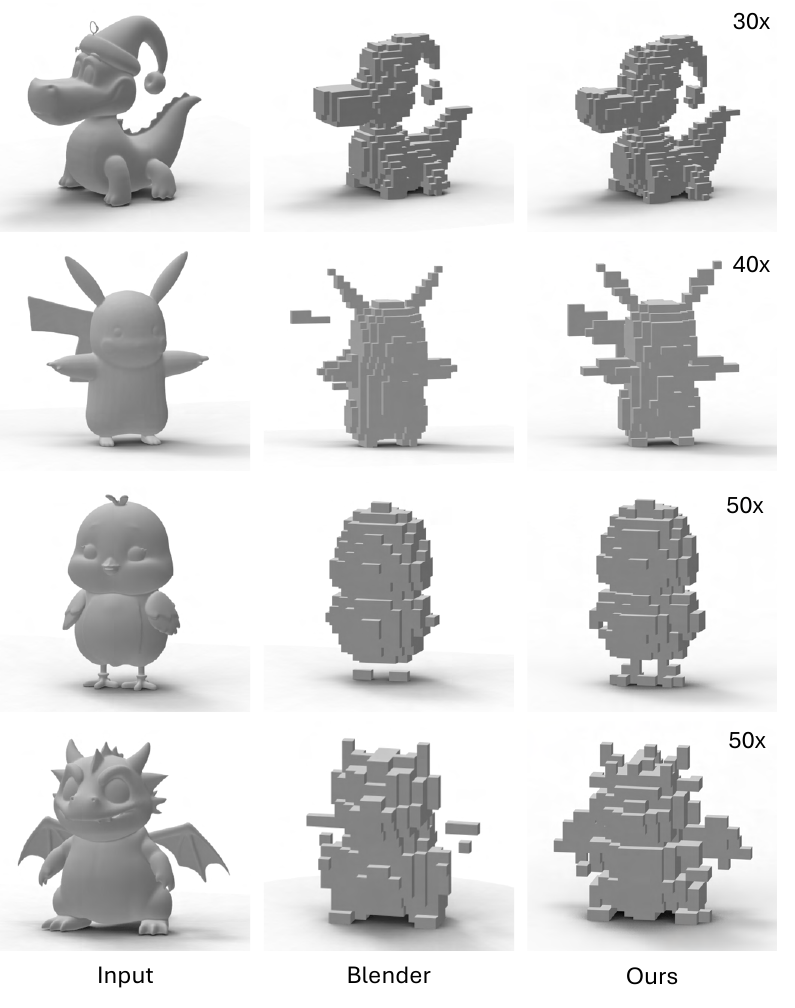}
    \caption{\textbf{Greyscale examples.}}
    \label{fig:grey}
\end{figure}

We conducted a user study with 72 participants, who were presented with \textbf{35 colored voxel art examples} and \textbf{4 grayscale voxel renderings} Fig.~\ref{fig:grey}. The interface is illustrated in Fig.~\ref{fig:userstudy}. 

Each colored example was accompanied by the following two questions:  

\begin{itemize}[leftmargin=12pt]
    \item \textbf{Abstract detail:}  
    ``Which voxel art version most clearly and prominently represents abstract details, such as facial features, clothing, and textures?''
    \item \textbf{Voxel art appeal:}  
    ``Which version looks most visually appealing as a voxel art character, like something you might see in Minecraft or a stylized game?''
\end{itemize}

For the grayscale examples, participants answered:

\begin{itemize}[leftmargin=12pt]
    \item \textbf{Geometry preservation:}  
    ``Which grayscale voxel rendering more closely resembles the original 3D mesh in terms of overall geometry?''
\end{itemize}

\begin{figure*}[t]
\centering
\includegraphics[width=\textwidth]{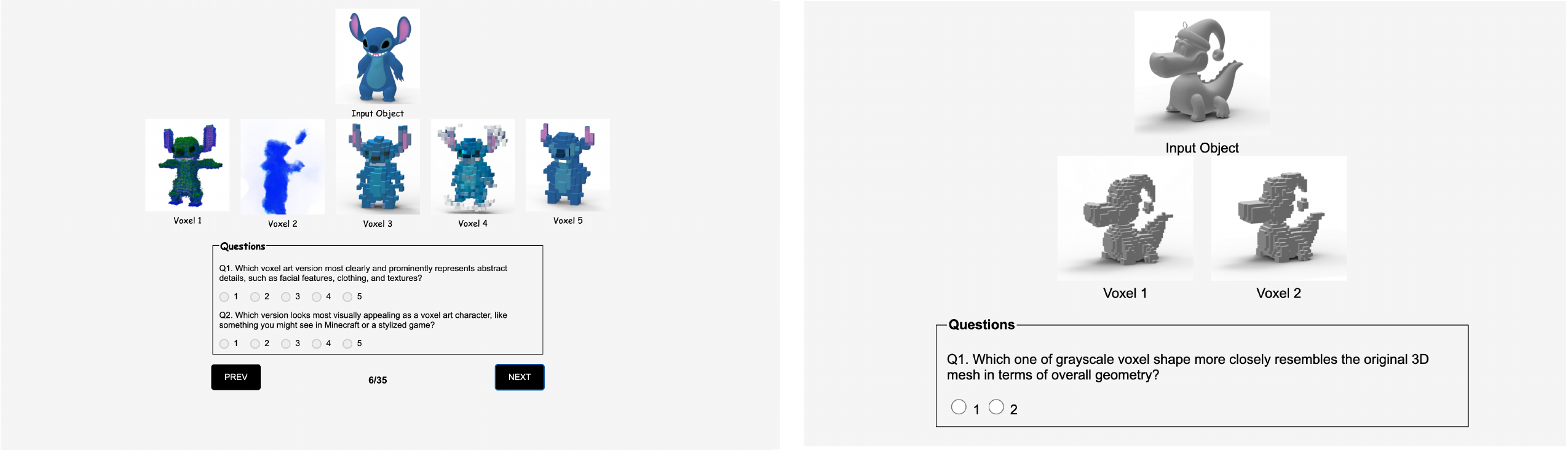}
\caption{\textbf{User study UI.}}
\label{fig:userstudy}
\end{figure*}

\noindent {\bf Expert study on color preference.}  
We further conducted a focused evaluation on color quantization with \textbf{10 art-trained participants}, all of whom had formal undergraduate education in art or design.  
Participants were asked to compare voxel art results with and without Gumbel-Softmax across \textbf{10 example pairs}, and answered the following two questions:  

\begin{itemize}[leftmargin=12pt]
    \item \textbf{Abstract detail:}  
    ``Which voxel art version most clearly and prominently represents abstract details, such as facial features, clothing, and textures?''
    \item \textbf{Voxel art appeal:}  
    ``Which version looks most visually appealing as a voxel art character, like something you might see in Minecraft or a stylized game?''
\end{itemize}

\begin{figure}[t]
\centering
\includegraphics[width=\columnwidth]{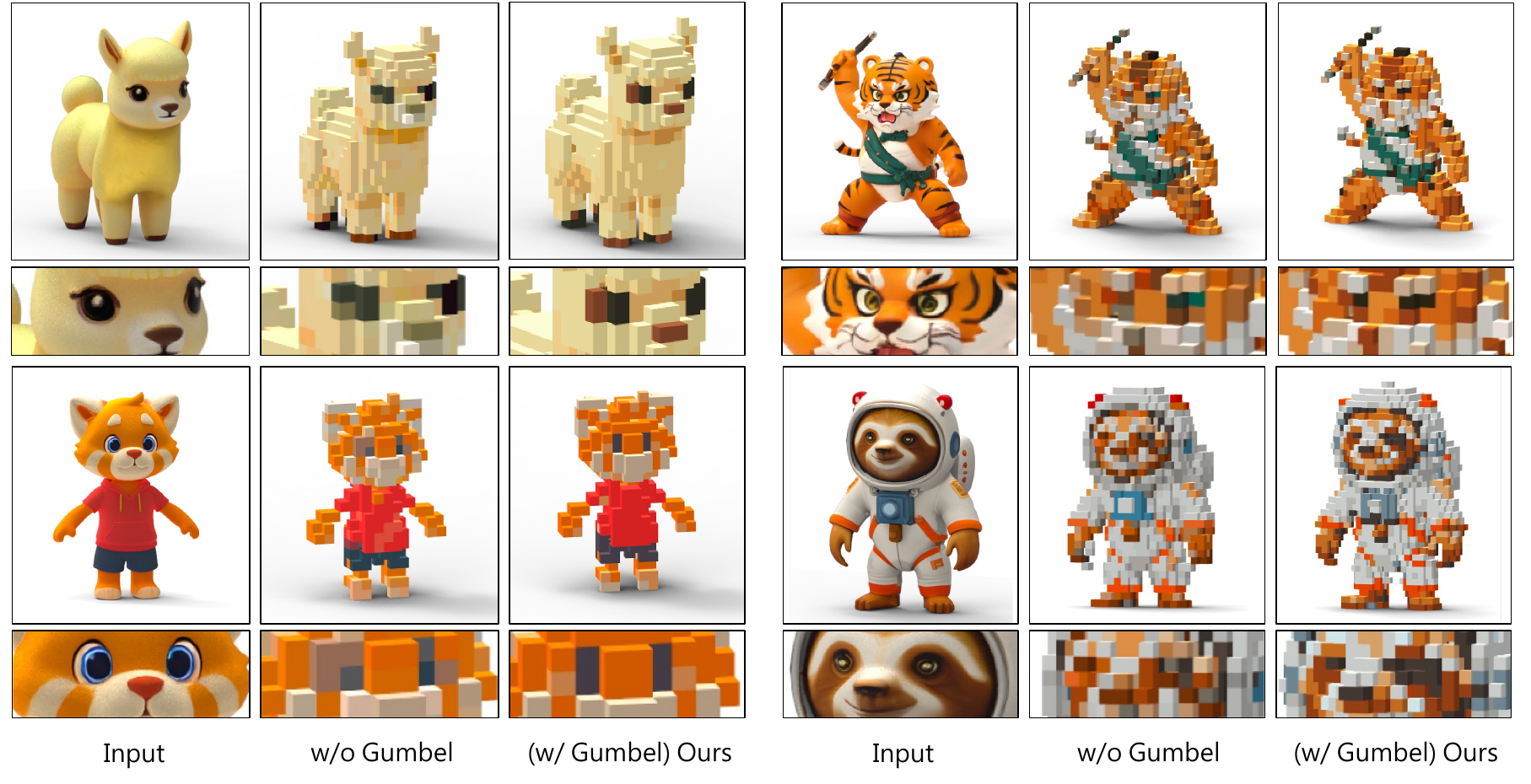}
\caption{\textbf{Ablation user study of Gumbel.} Four representative examples comparing results with and without Gumbel-Softmax. Without Gumbel-Softmax, voxel colors become blurred and features less distinct.}
\label{fig:gumbel_study}
\end{figure}

As illustrated in Fig.~\ref{fig:gumbel_study}, we present four representative examples comparing results \textit{with} and \textit{without} Gumbel-Softmax. Across responses from 10 participants on 10 question pairs, \textbf{88.89\%} favored the \textit{with Gumbel-Softmax} results for voxel-art appeal (Table~\ref{tab:wrap-user} (b)), confirming its importance in producing dominant tones and clear edges.

\vspace{3pt}
\noindent {\bf Runtime analysis.}
On a single RTX~4090, the total runtime of our two-stage pipeline is approximately $\sim$20 minutes, depending on the voxel resolution.
Overall, our method is substantially faster than manual voxel art creation.

\section{Additional Qualitative Results}
\label{sec:additional_results}

\noindent {\bf More comparisons with baselines.}  
In total, we evaluated \textbf{35} character models for CLIP-IQA.  
Here, we additionally present \textbf{8 representative examples} for qualitative comparison against the baselines: Pixel art to 3D extension, IN2N~\cite{haque2023instruct}, Vox-E~\cite{sella2023vox}, and Blender Geometry Nodes, as illustrated in Fig.~\ref{fig:appendix_baselines}. While IN2N~\cite{haque2023instruct} is effective in certain cases, we found it often fails in our setting.  
This is mainly because each guidance image used during training can differ significantly, leading to large inconsistencies across views.  

\begin{figure*}[t]
\centering
\includegraphics[width=\textwidth]{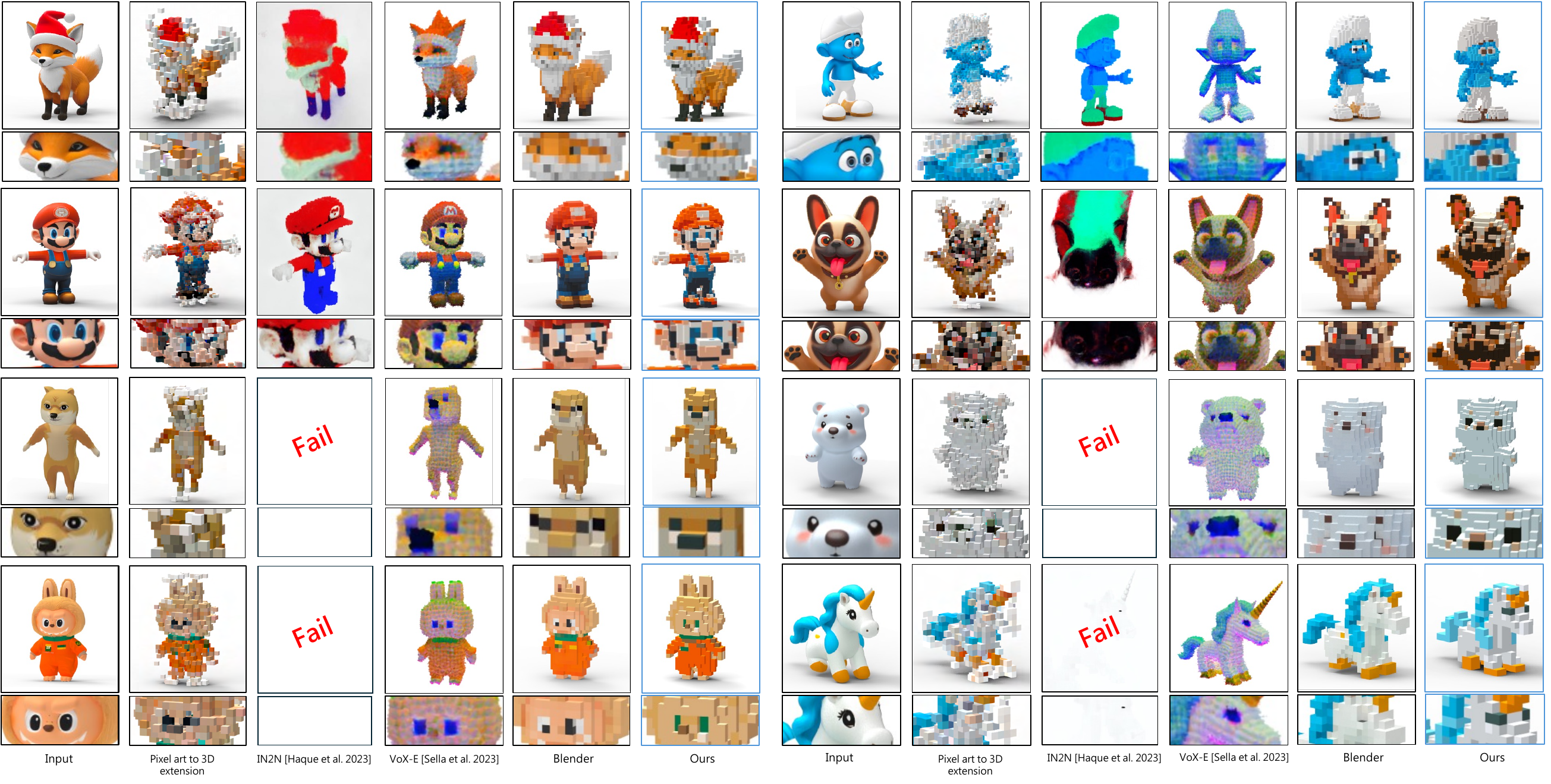}
\vspace{-6mm}
\caption{\textbf{Additional qualitative comparisons with baselines.} 
Eight representative examples compared with Pixel, IN2N, Vox-E, and Blender Geometry Nodes.}
\label{fig:appendix_baselines}
\end{figure*}

\noindent {\bf Results with varying palette settings.}
As shown in Fig.~\ref{fig:appendix_palette}, we present comparisons under different color selection strategies and palette sizes, with K-means adopted as our default palette extraction method.

\noindent {\bf Results under different voxel sizes.}
Fig.~\ref{fig:appendix_resolution} illustrates voxel art renderings generated with varying voxel resolutions, demonstrating how grid granularity influences the level of abstraction, sharpness of edges, and overall visual fidelity of the outputs.

\begin{figure}[t]
\centering
\includegraphics[width=\linewidth]{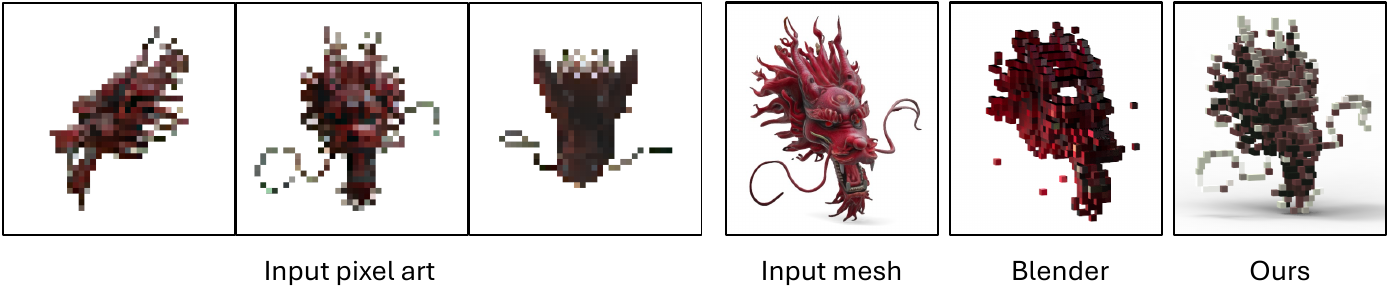}
\caption{\textbf{Representative failure cases.} Complex shapes with fine-grained geometric details are difficult to represent under limited voxel resolution, resulting in loss of intricate structures.}
\label{fig:appendix_failure}
\end{figure}

\begin{figure*}[h]
\centering
\includegraphics[width=\textwidth]{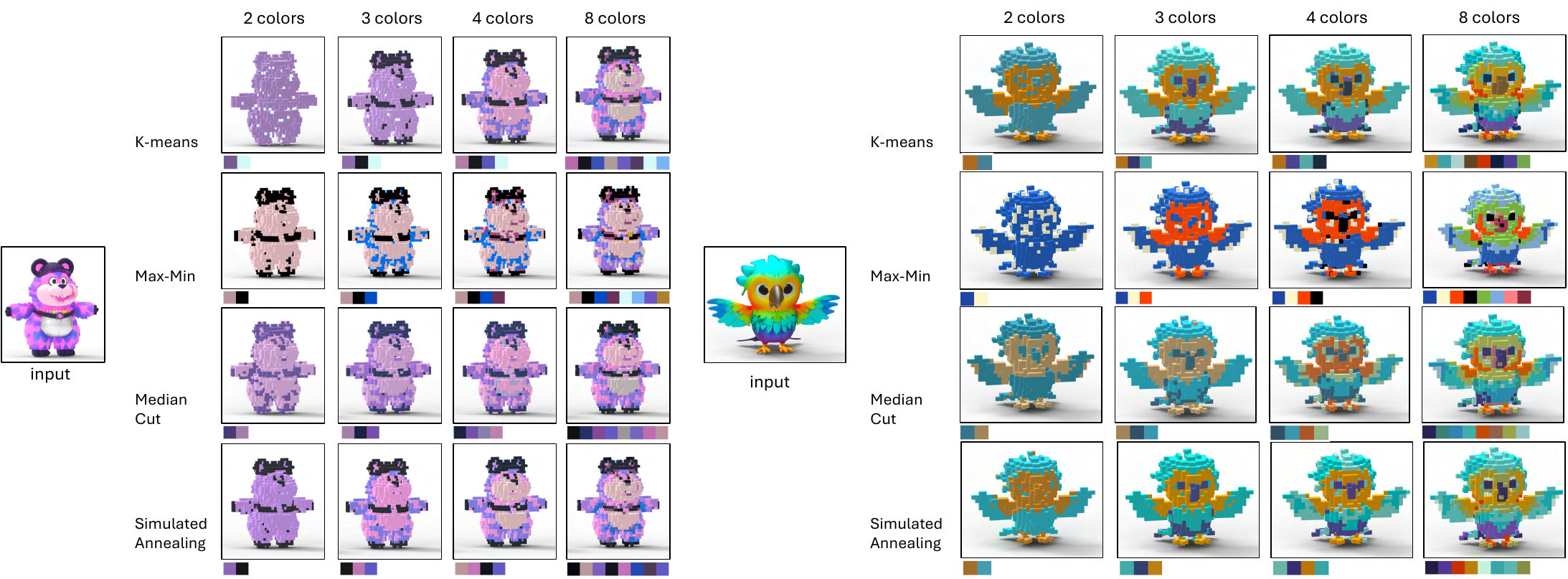}
\caption{\textbf{Results with varying palette settings.} 
Examples using different palette extraction strategies and palette sizes.}
\label{fig:appendix_palette}
\end{figure*}

\begin{figure*}[h]
\centering
\includegraphics[width=\textwidth]{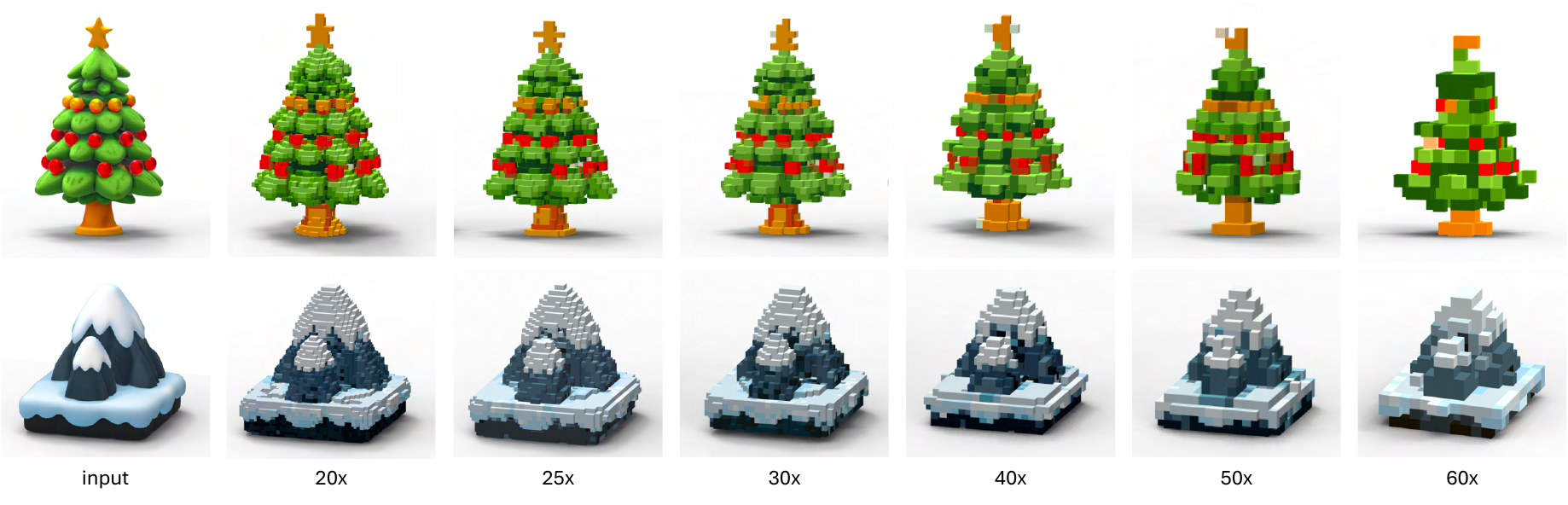}
\caption{\textbf{Results under different voxel sizes.}}
\label{fig:appendix_resolution}
\end{figure*}

\begin{figure*}[t]
    \centering
    \includegraphics[width=\textwidth]{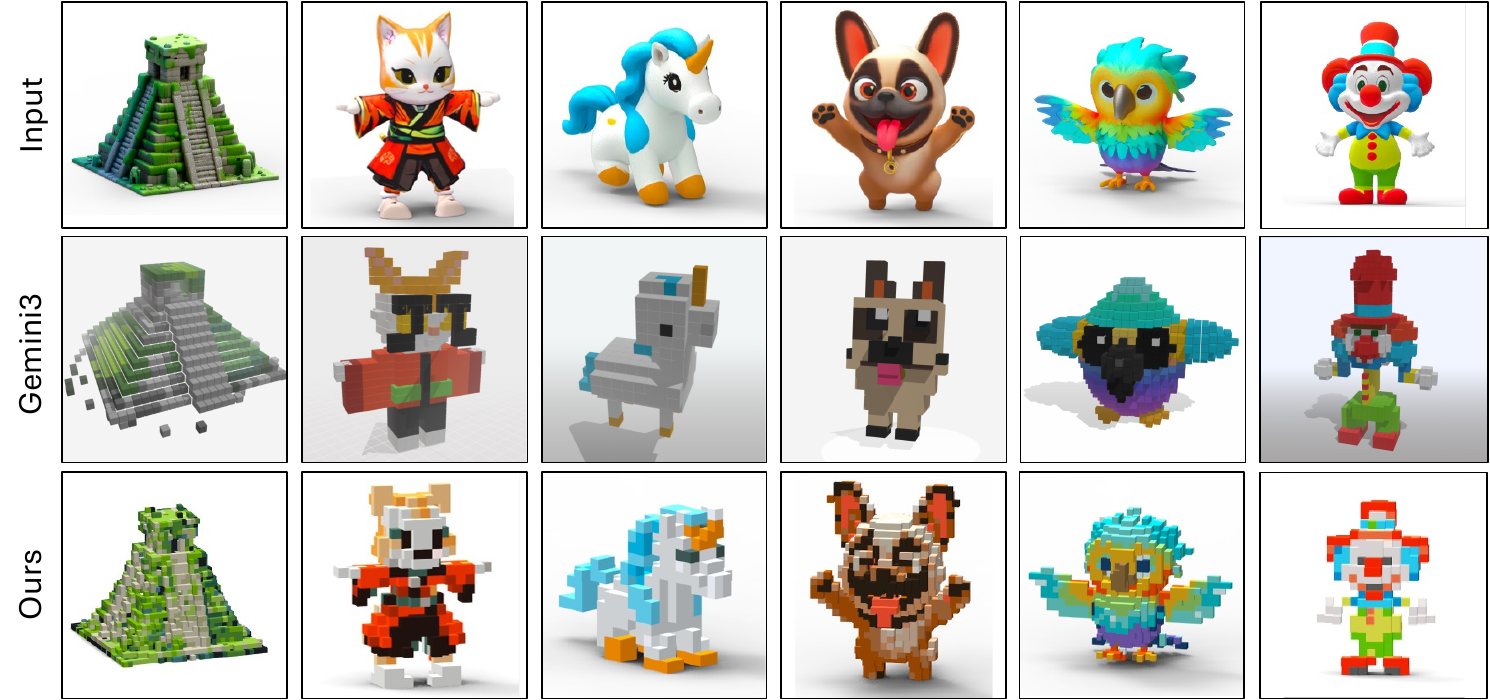}
    \caption{\textbf{Comparison with Gemini 3}~\cite{google_gemini}. 
    While Gemini 3 can generate voxel art through code, it lacks precise control over resolution, palette, and visual fidelity to input references.}
    \label{fig:Gemini3}
\end{figure*}

\begin{figure*}[t]
\centering
\includegraphics[width=\linewidth]{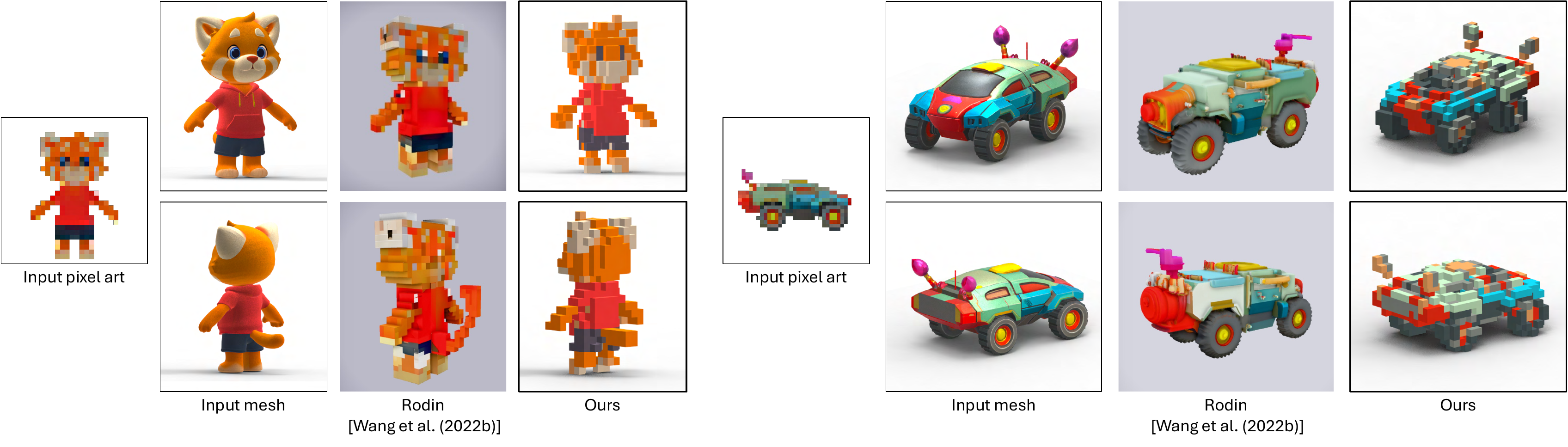}
\caption{\textbf{Comparison with Rodin}~\cite{wang2022rodin}. 
Rodin excels at image-to-mesh but is not tailored for voxel art, often yielding non-voxel outputs (right) or flat geometry (left).}
\label{fig:Rodin}
\end{figure*}

\noindent {\bf Comparison with LLM-based voxel generation.}  
We also compare with \textbf{Gemini 3}~\cite{google_gemini}, the latest state-of-the-art large language model, which can generate 3D voxel art through code generation in AI Studio.
As shown in Fig.~\ref{fig:Gemini3}, while Gemini 3 excels at creating interactive voxel-based applications and can produce detailed voxel art through its advanced coding capabilities, it lacks precise control over abstraction details, resolution, and color palette selection.
In contrast, our method enables fine-grained control over voxel resolution and palette constraints while faithfully preserving the visual characteristics through multi-view optimization.
This demonstrates the advantage of Voxify3D for controllable and appearance-faithful voxel art generation.

\noindent {\bf Comparison with single-image 3D reconstruction.}  
We also compare with \textbf{Rodin}~\cite{wang2022rodin}, which performs well for image-to-mesh generation but is not designed for voxel art.  
As shown in Fig.~\ref{fig:Rodin}, Rodin sometimes produces non-voxel outputs (right), and due to the single-image input, it often fails to capture reliable depth, resulting in flat structures (left).  
This further underscores the benefit of our multi-view voxel optimization pipeline.

\section{Failure cases and analysis}
\label{sec:failure_cases}
Finally, representative failure cases are shown in Fig.~\ref{fig:appendix_failure}, mainly arising from complex shapes that exceed the capacity of the limited voxel resolution. These examples suggest that voxel art is better suited for capturing abstract details conveyed through color patterns and tonal contrasts, whereas fine-grained geometric intricacies are more likely to be lost under coarse discretization. A promising future direction is to adopt adaptive voxel resolutions, where regions requiring fine details use smaller voxels while simpler areas maintain coarser ones, enabling better preservation of geometric complexity without sacrificing the aesthetic appeal of voxel art.

 \fi

\end{document}